\documentclass[letterpaper, 10 pt, conference]{ieeeconf}  

\IEEEoverridecommandlockouts                              

\usepackage{cite}
\usepackage{amsmath,amssymb,amsfonts, mathtools}
\usepackage{algorithmic}
\usepackage{graphicx}
\usepackage{textcomp}
\usepackage{xcolor}
\usepackage{subcaption}
\usepackage{multirow}
\usepackage{comment}
\usepackage{todonotes}
\usepackage{url}
\usepackage{threeparttable, tablefootnote}
\usepackage[font=small]{caption}

\def\BibTeX{{\rm B\kern-.05em{\sc i\kern-.025em b}\kern-.08em
    T\kern-.1667em\lower.7ex\hbox{E}\kern-.125emX}}

\title{\LARGE \bf
 Real-Time Context-aware Detection of \\ Unsafe Events in Robot-Assisted Surgery}
 
\author{Mohammad Samin Yasar, Homa Alemzadeh
\thanks{*This work was partially supported by a grant from the U.S. National Science Foundation (Award No. 1748737).}
\thanks{*Authors are with the Department of Electrical and Computer Engineering (ECE), University of Virginia, Charlottesville, VA 22904, USA 
      {\tt\small\{msy9an, ha4d\}@virginia.edu}.}%
}
\begin{document}

\maketitle


\begin{abstract}
Cyber-physical systems for robotic surgery have enabled minimally invasive procedures with increased precision and shorter hospitalization. However, with increasing complexity and connectivity of software and major involvement of human operators in the supervision of surgical robots, there remain significant challenges in ensuring patient safety. This paper presents a safety monitoring system that, given the knowledge of the surgical task being performed by the surgeon, can detect safety-critical events in real-time. Our approach integrates a surgical gesture classifier that infers the operational context from the time-series kinematics data of the robot with a library of erroneous gesture classifiers that given a surgical gesture can detect unsafe events. Our experiments using data from two surgical platforms show that the proposed system can detect unsafe events caused by accidental or malicious faults within an average reaction time window of 1,693 milliseconds and F1 score of 0.88 and human errors within an average reaction time window of 57 milliseconds and F1 score of 0.76.

\end{abstract}
\section{INTRODUCTION}

\begin {figure*}[t!]
\centering
    \begin{subfigure}[t]{0.29\textwidth}
    \centering
    \includegraphics[width=1.\linewidth]{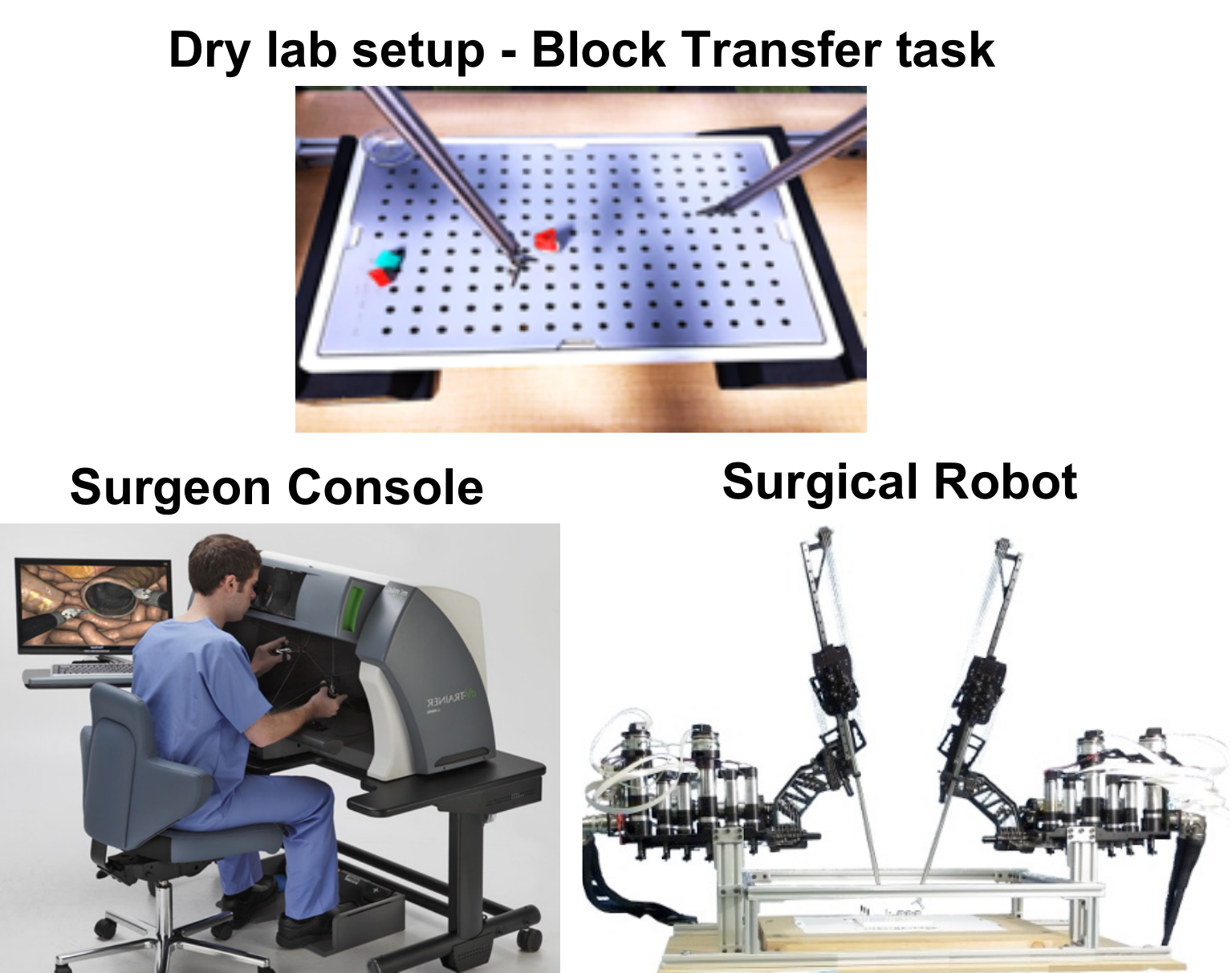}
    \subcaption{\centering}
    \label{fig:surgeon's console}
    \end{subfigure}
    \hfill
    \begin{subfigure}[t]{0.35\textwidth}
    \centering
    \includegraphics[width=1.14\linewidth]{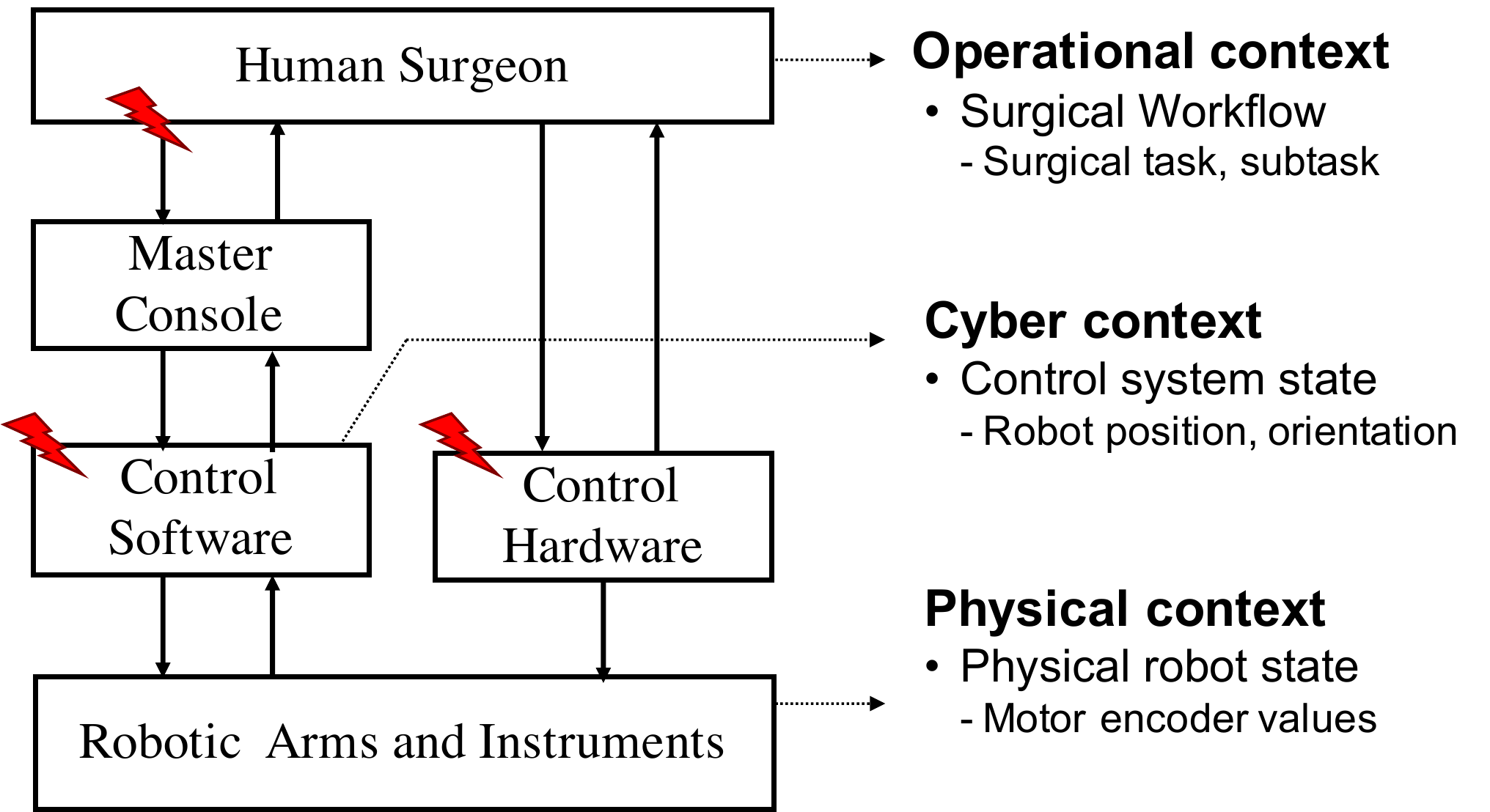}
    \subcaption{\centering}
    \label{fig:operational_context}
    \end{subfigure}
    \hfill
    \begin{subfigure}[t]{0.28\textwidth}
    \centering
    \includegraphics[width=0.9\linewidth, trim=7 8 10 10, clip]{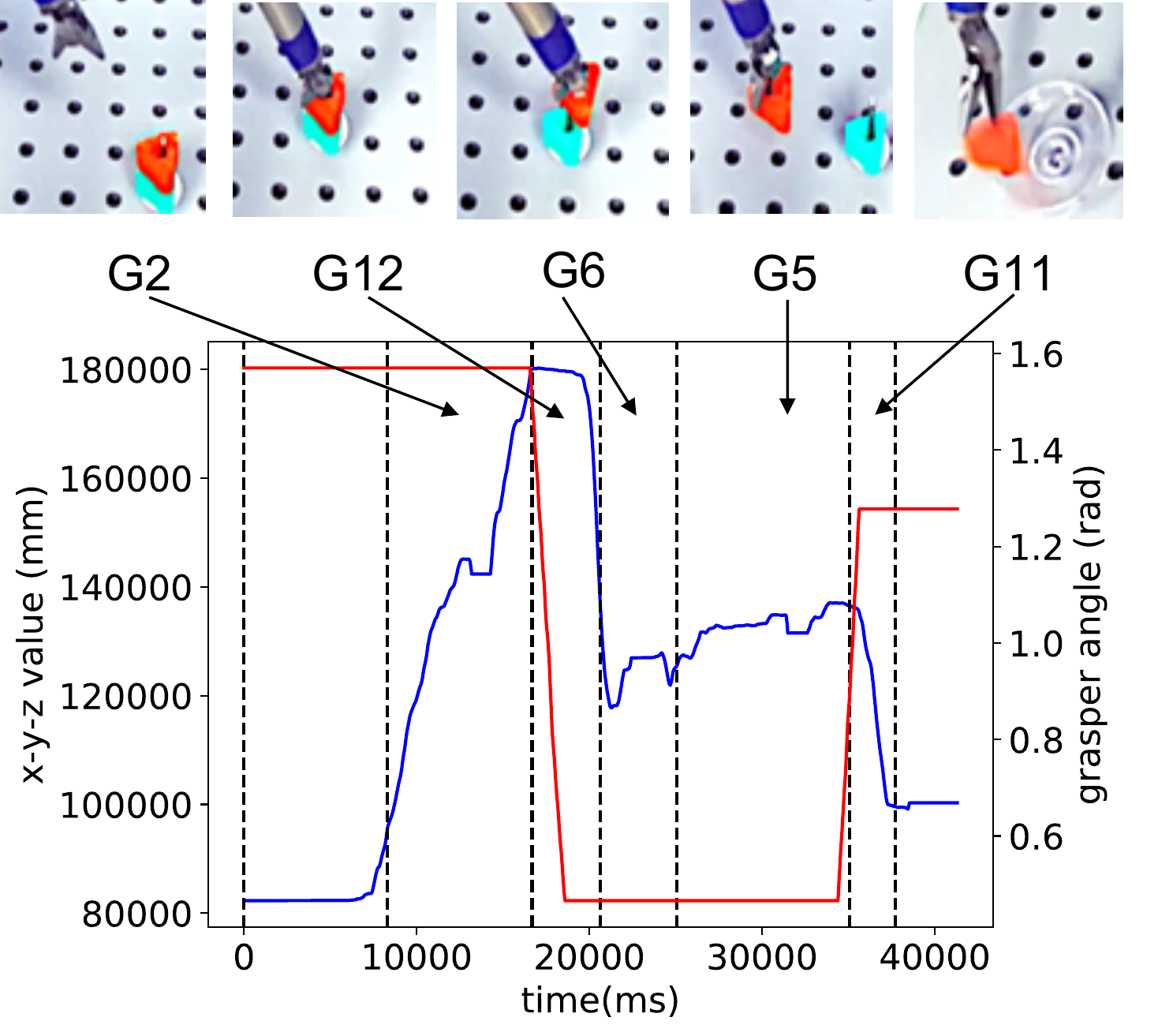}
    \subcaption{\centering}
    \label{fig:blocktransferwithcontext}
    \end{subfigure}
    \vspace{-0.5em}
    \caption{(a) Tele-operated surgical robot, adopted from \cite{hannaford2012raven, goldberg2012ergonomic}, (b) Typical control structure and system context in robotic surgery, (c) Operational context characterized by surgical gestures in a sample trajectory of Block Transfer task}
    \label{Fig:Overall Systems}
    \vspace{-1em}
\end{figure*}

Robot-assisted surgery (RAS) is now a standard procedure across various surgical specialties, including gynecology, urology and general surgeries. During the last two decades, over 2 million procedures were performed using the Intuitive Surgical's daVinci robot in the U.S.~\cite{Intuitive2017}. Surgical robots are complex human-in-the-loop Cyber-Physical Systems (CPS) that enable 3D visualization of surgical field and more precise manipulation of surgical instruments such as scissors, graspers, and electro-cautery inside patient's body. The current generation of surgical robots are not fully autonomous yet. They are in level 0 of autonomy \cite{yang2017medical}, following the commands provided by the surgeons from a master-side tele-operation console in real-time (Figure \ref{fig:surgeon's console}), translating them into precise movements of robotic arms and instruments, while scaling surgeon's motions and filtering out hand-tremors.  
By increasing flexibility and precision, surgical robots have enabled new types of surgical procedures and  have reduced complication rates and procedure times. 

Recent studies have shown that safety in robotic surgery may be compromised by vulnerabilities of the surgical robots to accidental or maliciously-crafted faults in the cyber or physical layers of the control system or human errors \cite{alemzadeh2016adverse,rajih2017error}. Examples of system faults include disruptions of the communication between the surgeon console and the robot, causing packet drops or delays in tele-operation~\cite{bonaci2015make}, accidental or malicious faults targeting the robot control software~\cite{alemzadeh2016targeted}, or faulty sensors and actuators~\cite{alemzadeh2016adverse} (Figure \ref{fig:operational_context}). Those faults can propagate and manifest as system errors (e.g., unintended movements, modification of surgeon's intent, and unresponsive robotic system \cite{alemzadeh2014systems}) or cause human errors (e.g., multiple attempts to suture or end-effector going out of sight \cite{eubanks1999objective, elhage2015assessment, joice1998errors}) during surgery and eventually lead to safety-critical events that negatively impact patients and caregivers. Examples include causing unexpected cuts, bleeding, or minor injuries or resulting in long procedure times and other complications during the procedure or afterwards~\cite{alemzadeh2016adverse}. Table \ref{table:common_errors} provides examples of common types of human errors during a sample set of surgical tasks, as reported in the literature. 

Our goal is to improve the safety of surgery by enhancing the robot controller with capabilities for real-time detection of early signs of adverse events and preventing them by issuing timely warnings or taking mitigation actions. Previous works on safety and security monitoring and anomaly detection in surgical robots and other CPS have mainly focused on incorporating the information from the cyber and/or physical layers for modeling and inference of the system context \cite{leveson2011engineering} and distinguishing between safe and unsafe control commands that could potentially lead to safety hazards. For example,~\cite{alemzadeh2016targeted} proposed an anomaly detection framework for detection of targeted attacks on the robot control system through modeling robot physical state and dynamics. They showed that considering the physical context (e.g., next motor position and velocities to be executed on the robot) leads to improved detection of unsafe events compared to fixed safety checks on just the cyber state variables (e.g., torque commands calculated in control software). However, with the major involvement of the human operators in real-time control and operation of semi-autonomous CPS such as surgical robots, another important contributing factor to safety is the operational context that captures human operators' preferences, intent, and workflow. 
In this work, we aim to improve the detection coverage for unsafe events during surgery by considering a more complete view of system context that incorporates the knowledge of the surgical workflow, characterized by the surgical tasks or gestures being performed by the surgeon. This is motivated by our preliminary results in~\cite{yasar2019context} that showed encouraging evidence for use of gesture-specific safety constraints in early detection of safety-critical events.

This paper presents an online safety monitoring system for robot-assisted surgery that takes the human operator actions (surgeon's commands at the tele-operation console) as input to infer the operational context (current surgical subtask or gesture) and performs context-specific validation of the kinematics state of the robot to detect 
erroneous gestures that could potentially lead to safety-critical events. The proposed system extends our early design of a context-aware monitor in~\cite{yasar2019context} and can be integrated with the existing surgical robots and simulators to provide real-time feedback to surgeons and potentially improve the learning curves in simulation training and prevent adverse events in actual surgery. 
In summary, the main contributions of the paper are as follows:
\begin{itemize}
    \item Demonstrating that across a surgical task, the errors are context-specific, i.e., dependent on the current surgical gesture being performed by the surgeon (Section \ref{subsection:distibution analysis}). This suggests the possibility of designing a context-aware monitoring system that can detect gesture-specific errors and can be pervasively applied to any surgical task that is composed of such atomic gestures. 
    
    \item Developing a safety monitoring system, consisting of a surgical gesture classifier and a library of gesture-specific classifiers that can detect the erroneous gestures in real-time (Section \ref{section: methods}).
    Our proposed system can detect errors caused by accidental faults or attacks targeting the software, network or hardware layer or human errors that affect the kinematic state of the robot. 

    \item Introducing a simulation environment based on ROS \cite{quigley2009ros} and Gazebo \cite{koenig2004design} and the Raven II~\cite{hannaford2012raven} control software (an open-source surgical robot from Applied Dexterity Inc.), that enables the experimental evaluation of safety monitoring solutions for robotic surgery (Section \ref{subsubection: Gazebo Simulator}). Our simulator can model physical interactions between the robot and its environment, generate synthetic surgical trajectory data, and simulate realistic robot failure modes using software fault injection without harming the physical robot.
     
    \item Evaluating our monitoring system in terms of accuracy and timeliness in detecting errors using data from two different surgical platforms, Raven II and daVinci Research Kit\cite{kazanzides2014open} (dVRK from Intuitive Surgical Inc.). Our results in Section \ref{section: results} provide evidence for our hypothesis that a context-aware safety monitor can enable more accurate detection of anomalies. The proposed monitor can detect potential adverse events for the surgical tasks of Block Transfer and Suturing with average F1 scores of 0.88 and 0.76 within average reaction time windows of 1,693 and 57  milliseconds, respectively. 
    
\end{itemize} 

\begin{table}[b!]
\begin{center}
\vspace{-1em}
\setlength\tabcolsep{5pt}
\resizebox{\linewidth}{!}{
\begin{tabular} {|c|c|c|c|}
 \hline
\textbf{Surgical Task} & \textbf{Faults} & \textbf{Errors} & \textbf{Adverse Events} \\
\hline
Laparoscopic & Wrong orientation& Liver laceration & Hematoma \cite{bonrath2013defining} \\
Cholecystectomy &of end-effector  & with bleeding  &
\\
\hline
Laparoscopic & Too much force& Peroration of & Subhepatic  \cite{bonrath2013defining} \\ 
Cholecystectomy &with grasper  & gallbladder wall  & abcess
\\
\hline
Anastomosis & Wrong Cartesian  & Needle   & Punctures \cite{bonrath2013error} \\
 & Position & overshoots goal  & Vessel \\ 
\hline
\end{tabular}}
\caption{Common Errors in Surgery}
\label{table:common_errors}

\end{center}
\end{table}

\section{Preliminaries}
\label{subsection: methods}
Our goal is to enhance the surgical robots with capabilities for real-time detection of errors 
and providing just-in-time feedback to surgeons or taking automated mitigation actions before safety-critical events occur. 
Our safety monitoring solution is built upon the main concepts described next. \par

\par

\textbf{Operational Context in Surgery:}
\label{subsection:distibution analysis}
The diverse sources of faults and involvement of humans in the control of surgical robots make real-time detection of adverse events particularly challenging and require understanding of the surgical context in order to decide on the best safety actions to take. 

\begin {figure}[t!]
\vspace{-1em}
\begin{center}
    \includegraphics[width=0.6\linewidth]{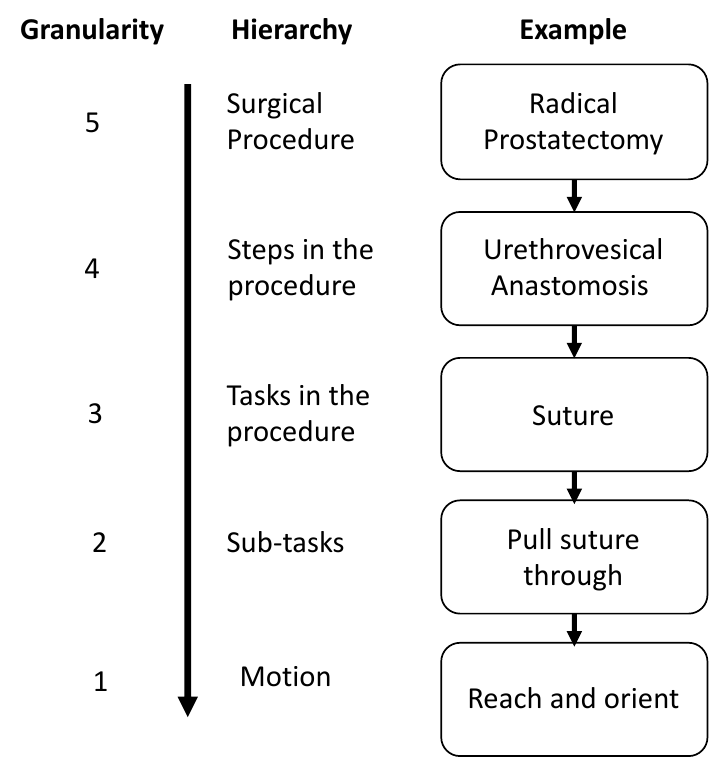}   
    \caption{\label{fig:hierarchy} Hierarchies in Surgical Procedures (adopted from \cite{neumuth2011modeling})}
\end{center}
\vspace{-2.5em}
\end{figure}

In our previous work\cite{yasar2019context}, we defined context in surgical procedures as a hierarchy, starting from the surgical procedure that is being executed, to the steps in the procedure, to surgical tasks, subtasks or gestures, and finally the specific motions of the robot (see Figure \ref{fig:hierarchy}). 
Within a specific procedure (e.g., Radical Prostatectomy) for a surgical task (e.g., suturing), the change in operational context happens in the temporal domain as a result of the change of the surgeon’s gestures or the position and orientation of the surgical instruments end-effectors, leading to the corresponding change in the subtask (e.g., pull suture through). Other contributing factors to the change in the operational context in surgery are the state of the robot software, the type of surgical instrument used, and the anatomical structures and their interactions within the surgical workspace. The operational context can be either observed directly using video data or inferred from the corresponding kinematics data and other state information from the robot.  \par 

\begin{figure} [t]

    \captionsetup[subfigure]{justification=centering}
    \centering
     \begin{subfigure}[b]{0.4\textwidth}
    \centering
    \includegraphics[width=0.845\linewidth ]{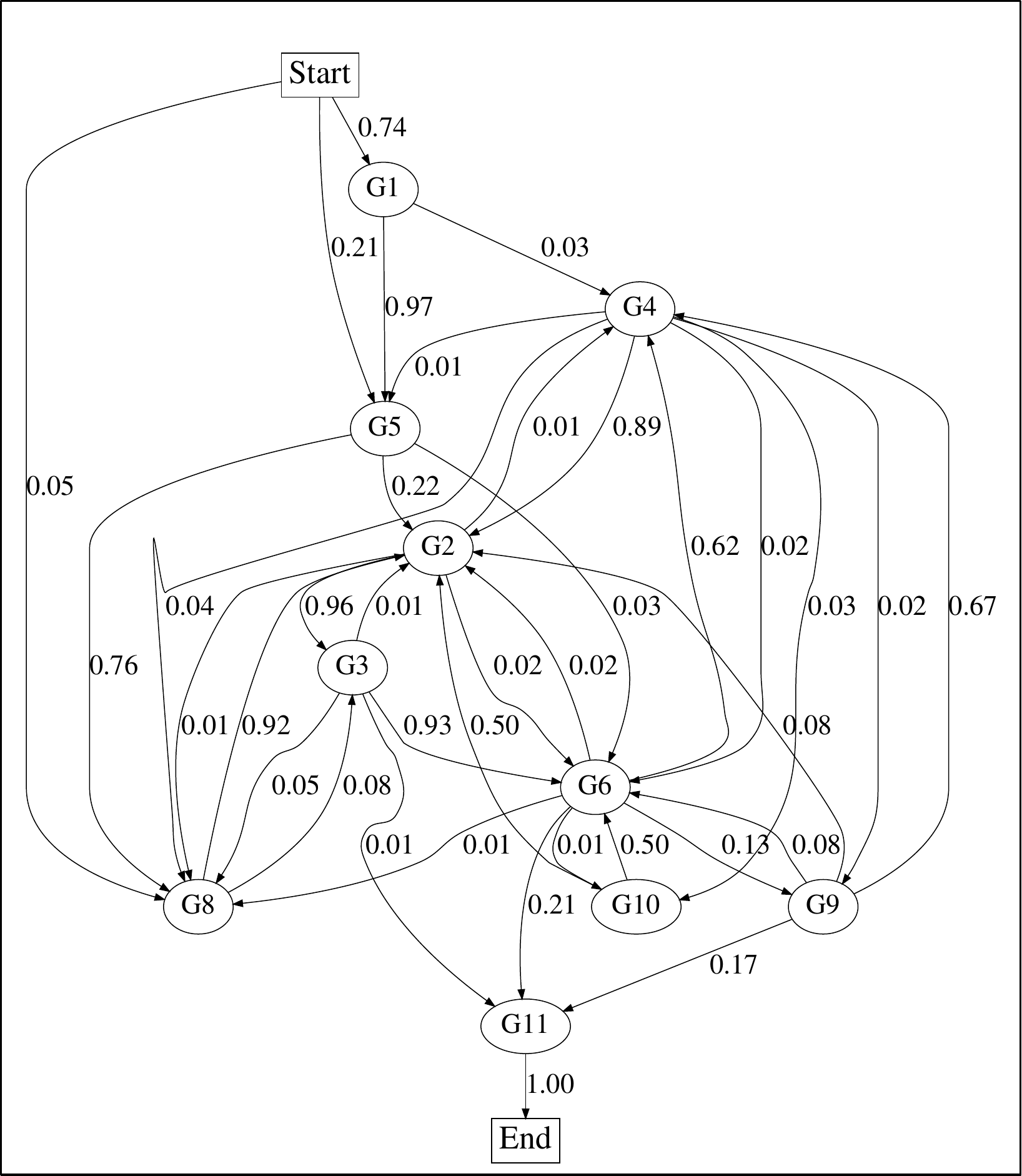}
    \subcaption{\centering Suturing}
    \label{fig:cfg_suturing}
    \end{subfigure}%
    \begin{subfigure}[b]{0.1\textwidth}
    \centering
    \includegraphics[width=0.815\linewidth]{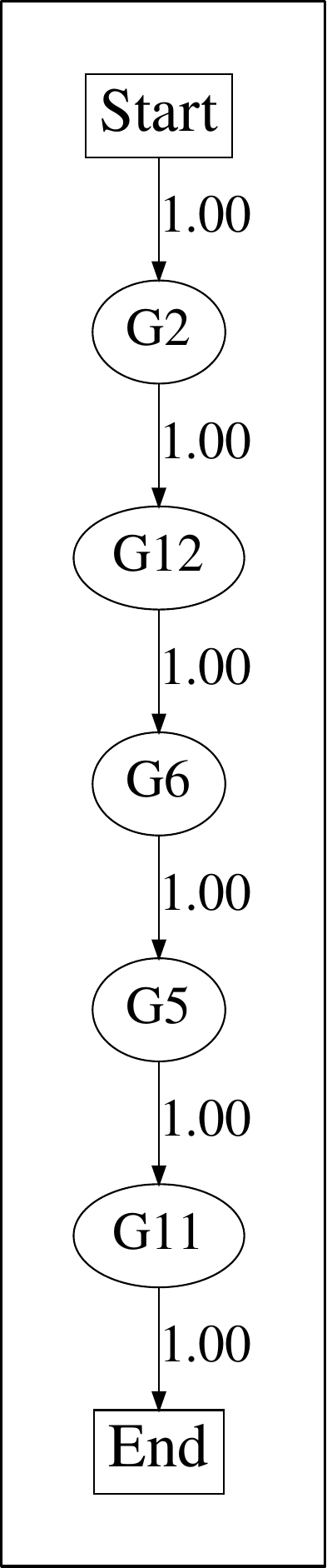}
    \subcaption{\centering Block Transfer}
    \label{fig:cfg_blocktransfer}
    \end{subfigure}
    \caption{Markov chain derived for the surgical task of Suturing and Block Transfer}
    \label{Fig:Markov Chains}
\vspace{-2em}
\end{figure}

One common approach to modeling of surgical tasks is using finite-state Markov chains with each state corresponding to an atomic surgical gesture\cite{rosen2006generalized}. In our work, we consider tasks from the Fundamentals of Laparoscopic Surgery (FLS) \cite{peters2004development}, in particular, Suturing and Block Transfer, which are commonly observed in both simulation training and actual surgery. Figure \ref{fig:cfg_suturing} shows the Markov chain representation of the task of Suturing, which we derived from the analysis of the dry-lab demonstrations in the JIGSAWS dataset \cite{gao2014jhu}. The gestures in Suturing are represented as G1 to G11, excluding G7, as described in Table \ref{tab:Rubric}. It is apparent that different demonstrations of Suturing could follow different sequences of gestures due to variations in surgeons' styles and common errors in performing the tasks. Figure \ref{fig:blocktransferwithcontext} shows a sample trajectory for the surgical task of Block Transfer, consisting of G2, G5, G6, G11 and G12 gestures. As this is a comparatively simple task, all demonstrations in our collected dataset have the same sequence as seen by the Markov chain in Figure \ref{fig:cfg_blocktransfer}, making the probability of transitioning between different states 1. \par

Prior works on surgical skill evaluation \cite{reiley2009decomposition} and our preliminary work on safety monitoring in surgery~\cite{yasar2019context} have shown that efficiency and safety of surgical tasks are context-specific and that certain gestures or sub-tasks are better indicators of surgeon's skills and surgical outcomes \cite{zia2018surgical, hung2018utilizing, zia2019novel}. We incorporate this concept into our system and provide evidence of improved results if we use the notion of gestures when detecting unsafe events.\par

Current surgical robots and simulators use surgeon's commands and robot trajectories, collected from surgical procedures or virtual training experiments, for \textit{offline} analysis of subtasks and objective evaluation of surgeon's performance~\cite{rosen2006generalized, reiley2009decomposition, ahmidi2015automated, fard2018automated, fawaz2018evaluating}. 
In this work, we show that there is potential for the \emph{online} analysis of this data during surgery to prevent the occurrence of safety-critical events.

\begin{table}[b!]
    \centering
    \vspace{-1.5em}
    \resizebox{\linewidth}{!}{
    \begin{tabular}{|p{0.09\linewidth}|p{0.25\linewidth}|p{0.25\linewidth}|p{0.2\linewidth}|p{0.21\linewidth}|}
    \hline
        Gesture & Description & \multicolumn{2}{c|}{Common Gesture Specific Errors} & Potential\\
        Index &  & \multicolumn{2}{c|}{(Failure Modes)} & Causes (Faults)\\
    \hline
        G1 & Reaching for needle with right hand & More than one attempt to reach &  & Wrong rotation angles \\
    \hline
        G2 & Positioning needle & More than one attempt to position & & Wrong rotation angles \\
    \hline
        G3 & Pushing needle through the tissue & Driving with more than one movement & Not removing the needle along its curve & Wrong Cartesian Position \\
    \hline
        G4 & Transferring needle from left to right & Unintentional Needle Drop & Needle held on needle holder not in view at all time & Wrong Cartesian Position/Sudden jumps \\
    \hline 
        G5 & Moving to center with needle in grip & Unintentional Needle Drop & & High Grasper Angle \\
    \hline
        G6 & Pulling suture with left hand & Needle held on needle holder not in view at all times &Unintentional Needle Drop & Wrong Cartesian Position/Sudden jumps \\
    \hline
        
        G8 & Orienting needle & Uses tissue/ instrument for stability & More than one attempt at orienting & Wrong rotation angles  \\
    \hline
        G9 & Using right hand to help tighten suture  & Knot left loose  &  & Low pressure applied to tighten suture \\
    \hline
    G10 & Loosening more suture  &  &  &\\
    \hline
        G11 & Dropping suture and moving to end points & Failure to dropoff &  & Low Grasper Angle  \\
    \hline
        G12 & Reaching for needle with left hand & More than one attempt to reach & & Wrong Cartesian Position/Sudden jumps  \\
    \hline
    \end{tabular}}
    \caption{Gesture specific errors in Suturing and Block Transfer tasks (adopted from \cite{gao2014jhu,moorthy2004bimodal})}
    \label{tab:Rubric}
\end{table} 

\textbf{Erroneous Surgical Gestures:} 
Given the knowledge of surgical gestures, the goal of the safety monitor is to detect erroneous gestures performed on the surgical robot that could indicate the early signs of unsafe events. 
As there are many variables involved each time a surgical task is performed, starting from the idiosyncrasies of the surgeon (preferences, efficiency, and expertise) to the dynamics of underlying cyber-physical system of the robot, it is safe to assume that there will be many variations of the same surgical gesture. However, it is imperative to identify erroneous gestures that could potentially lead to adverse events or delay in the task. The identification of erroneous gestures as the atomic building blocks of surgical procedures could enable preemptive detection of unsafe events in any surgical task.

We extend the well-established definition of surgical gestures by identifying the common errors that are observed when performing each gesture, using a rubric adopted from \cite{moorthy2004bimodal}. Table \ref{tab:Rubric} shows the set of gestures and common errors in the tasks of Suturing and Block Transfer. Similar rubrics can be derived for other tasks by identifying their atomic gestures and gesture-specific errors, as shown in~\cite{yasar2019context, elhage2015assessment, eubanks1999objective, joice1998errors, kwaan2006incidence}. We classify a gesture as erroneous if any of the common errors specific to that gesture are observed. Not all erroneous gestures will lead to adverse events. Depending on the gesture, the type of error, and other contextual factors, the erroneous gestures can vary in terms of severity but in this work we do not consider this for detecting them.  

We also show the types of faults in the kinematic state variables that can potentially cause such errors. We assume that accidental or malicious faults in software, hardware or network layer, or human errors can manifest as errors in the kinematics state variables and lead to such erroneous gestures. We demonstrate that this is possible through fault injection experiments on the RAVEN II surgical robot using simulated trajectory data for the task of Block Transfer in Section \ref{subsection: fault injection} and through analysis of the pre-collected trajectory data from dry-lab demonstrations of Suturing on the Intuitive Surgical's daVinci Research Kit in Section \ref{subsection: dvrk data}.

\begin{figure}[t!]
    \centering
    \includegraphics [width=1\columnwidth, trim=0 0 0 0, clip 
    ]{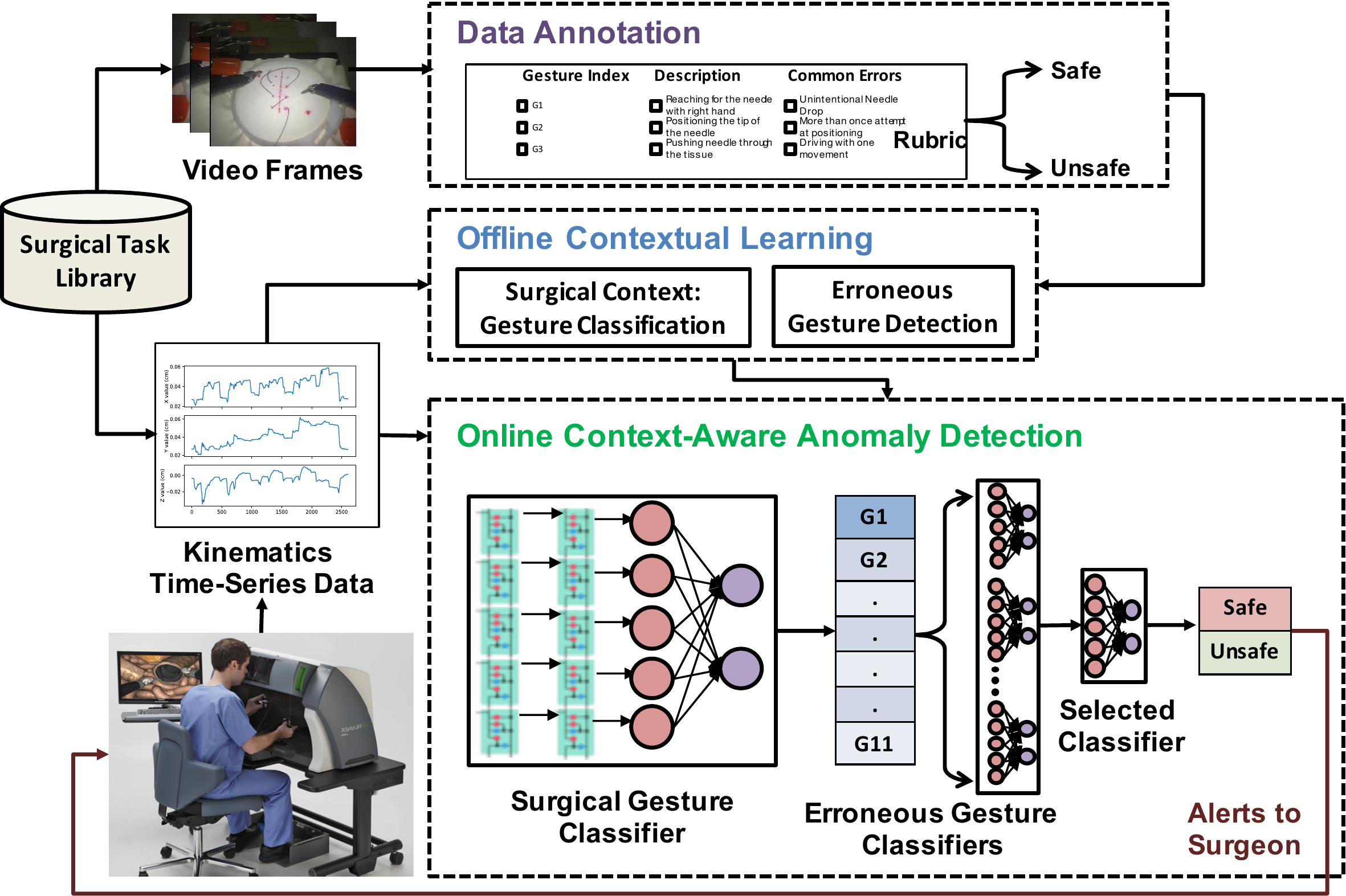}
    \caption{Pipeline for Real-time Context-aware Safety Monitoring System.} 
    \label{fig:current pipeline}
    \vspace{-2 em}
\end{figure}

\section{Context Aware Safety Monitoring}
\label{section: methods}
Our context-aware monitoring system
is composed of two supervised learning components, 1) A surgical gesture classifier followed by a library of 2) Erroneous gesture classifiers. 
 For both parts of our pipeline, we use variants of Deep Neural Networks (DNNs) \cite{lecun2015deep}, which have achieved state-of-the-art performance in many pattern-recognition and data mining problems. 
 We model the detection task as a hierarchical time-series classification problem. The first part of our pipeline is trained to identify the current operational context or gesture. This then activates the second part of the pipeline, which classifies the gesture as safe or unsafe by learning from gesture-specific spatio-temporal patterns in time-series kinematics data. We train the two parts of the pipeline separately from each other.
 Figure \ref{fig:current pipeline} shows our end-to-end pipeline for training and evaluation of the safety monitoring system\footnote{Code available at: https://github.com/UVA-DSA/ContextMonitor}. The proposed monitor can be integrated with the surgical robot by being deployed on a trusted computing base at the last computational stage in the robot control system pipeline~\cite{alemzadeh2016targeted} and be used in conjunction with other mechanisms proposed in previous works (see Section~\ref{section: related work}) to secure the robot against faults and attacks.\par 

\textbf{Analysis of Erroneous Gesture Distributions:}
\label{subsection: distribution analysis}
To better understand the characteristics of different erroneous gestures, we performed an analysis of their underlying distributions based on the kinematics data collected from the task of Suturing in the JIGSAWS database. Previous work \cite{krishnan2018transition, yasar2019context} have modeled the surgical trajectories as a multi-modal Gaussian distribution, with kinematics data being sampled from one of the many Gaussian mixtures and each mixture corresponding to a different gesture. We used this insight to estimate the probability density function of each erroneous gesture class using Gaussian kernels.  We then calculated the relative entropy between the respective  distributions of different erroneous gesture classes ($EG_i$) using the Jensen-Shannon Divergence (JS-divergence) metric \cite{lin1991divergence}  which provides us with a measure of difference between each pair of distributions as calculated in Equation \ref{equation:JS Divergence}:
\begin{equation}
    \begin{aligned}
    JSD (EG_i||EG_j) &= \frac{1}{2}D(EG_i||M) + \frac{1}{2}D(EG_j||M) \\
    where, \, M  &= \frac{1}{2}(EG_i+EG_j)  \, and  \\
    EG_{i,j} &= Erroneous \, Gestures
    \label{equation:JS Divergence}
    \end{aligned}
\end{equation}

\begin{figure}[t!]
    \centering
    \vspace{-1em}
    \centering
    \includegraphics[width=0.5\linewidth]{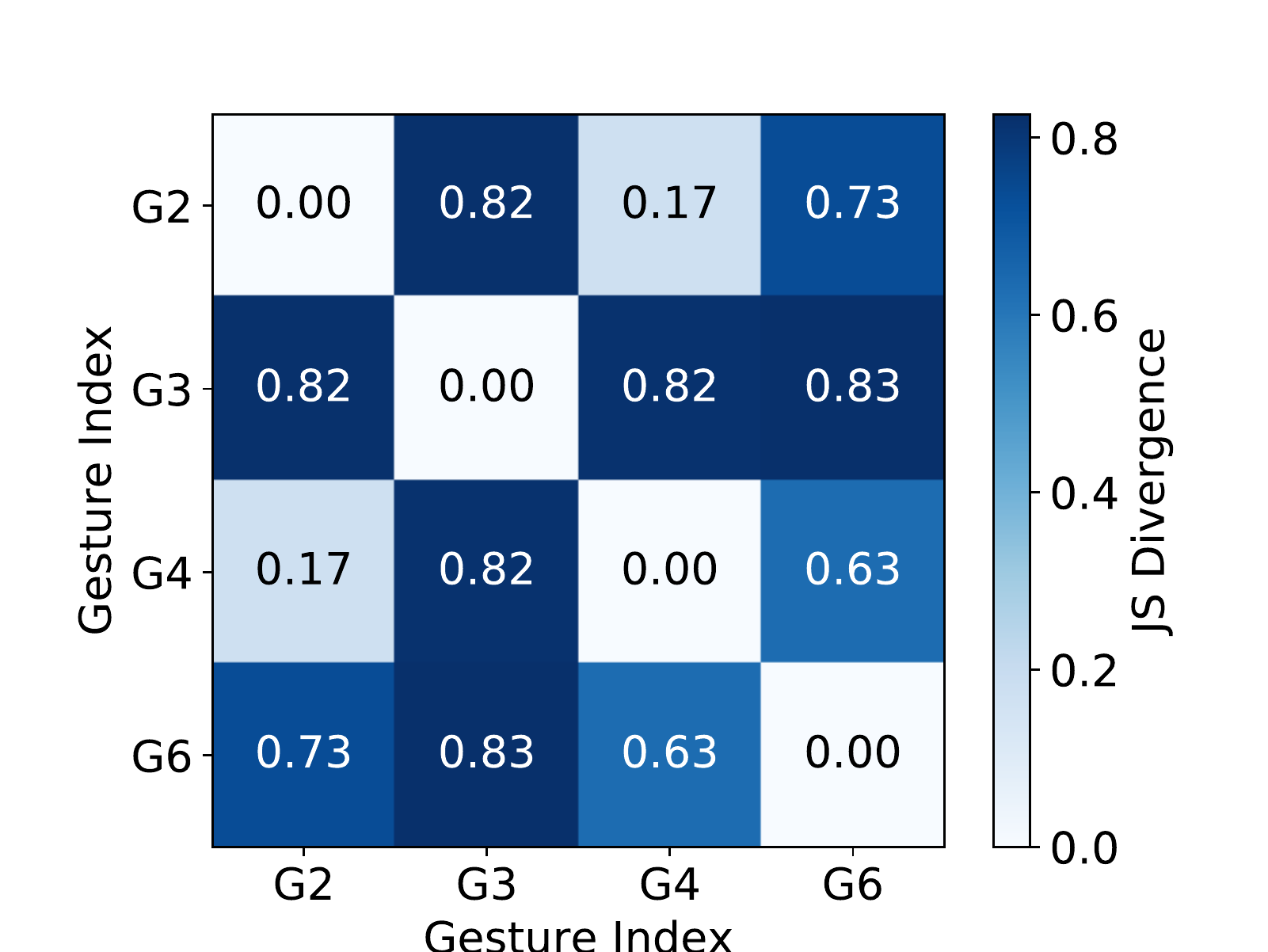}
    \label{fig:gesture-specific distributions}
    \caption{Pairwise divergence between erroneous gesture distributions}
    \label{fig:jd_siff}
    \vspace{-2em}
\end{figure}

\par
Figure \ref{fig:jd_siff} shows the pairwise JS-divergence between distributions of different erroneous gestures. We see that there is in particular a high divergence between the distributions of gesture classes G2, G3, G4 and G6, all of which are commonly occurring gestures and have a large number of samples in the task of Suturing (see Table \ref{tab:auc for each gesture}). For the other gesture classes we were not able to compute meaningful distributions due to small sample sizes. This observation partly supports our hypothesis that errors in surgery are context-specific and the knowledge of gestures might help with improved error detection. We use this observation in designing our \textit{Erroneous Gesture Detection} component of the pipeline by developing a library of classifiers, each trained for detecting errors in a specific gesture class.

\textbf{Gesture Segmentation and Classification:}
\label{subsection: gesture detection}
We model the task of identifying and segmenting surgical gestures as a multi-class classification problem (Equation \ref{equation: Classification Equation}). The input signal $x_t$ represents the time-series of kinematics variables with a sliding time-window of $w$ and a stride of $s$. The output $G_t$ represents the gesture corresponding to that time-series and is a one-hot vector of all gestures from 0 to 14. 
\begin{equation}
    \begin{array}{l}
    x_t = (x_t, x_{t+1},.., x_{t+w}) \\
    G_t = (0, 0, 1, .., 0)^T
    \label{equation: Classification Equation}
    \end{array}
\end{equation}

As the input signal is a multivariate time-series, we use Recurrent Neural Networks (RNN) which are known to learn spatial and temporal patterns. For our gesture classification, we use LSTM \cite{hochreiter1997long} networks which are known to learn long and short-term dependencies, and have the ability to decide what part of the previous output to keep or discard. 
A typical LSTM unit is composed of a memory cell and three gates: input, output and the forget gate. The gates regulate the flow of information inside the unit and allow LSTM architectures to remember information for long periods of time while also filtering information that is less relevant.

To aid our gesture classification and ensure smooth transition boundaries, we use stacked LSTM layers to provide greater abstraction of the input sequence and to allow the hidden states at each level to operate at a different timescale \cite{hermans2013training}. This is followed by a fully-connected layer with ReLU \cite{nair2010rectified} activation and a final softmax layer for obtaining gesture probabilities. The loss function is the categorical cross-entropy, with the model trained using the Adam \cite{kingma2014adam} optimizer. To address over-fitting, we use dropout regularization and early stopping on a held-out validation set. To improve the learning process, we use batch normalization layers and adaptive learning rate with step-decay.


\textbf{Erroneous Gesture Detection:}
\label{subsection: anomaly detection classifier}
Having identified the current gesture, $G_t$, the next stage of the pipeline classifies the gesture as erroneous or non-erroneous using the kinematics samples as input. We train this part of the pipeline separately from the gesture classification component and only combine the two parts in the evaluation phase, with the erroneous gesture detection following the gesture segmentation and classification (see Section \ref{section:evaluation of whole pipeline}). 
\begin{equation}
    \begin{array}{l}
    x_t = (x_t, x_{t+1},.., x_{t+w}) \\
    y_t = p(EG_t|G_t, x_t)
    
    \label{equation: Erroneous gesture equation}
    \end{array}
\end{equation}

We frame the problem as detection of a context-specific conditional event, i.e., a part of the trajectory can be erroneous or non-erroneous, depending on the current gesture, as shown in Equation \ref{equation: Erroneous gesture equation}. The input is the predicted gesture and a kinematics sample corresponding to that gesture, and the output is a binary classification of the kinematics sample 
to safe or unsafe. If any sample within a gesture is erroneous, we label that whole gesture as unsafe. 
Although our model is trained on sliding time-window samples instead of the whole gesture, it learns to have smooth output over time, allowing it to distinguish between entire boundaries of erroneous or non-erroneous gestures.\par
As a baseline, we trained a single classifier, with no explicit notion of context, for detecting the erroneous gestures. In this case, the problem reduces to a non-conditional binary classification of the time-series data, with the input being the kinematics sample and the output being whether it is erroneous or not. Similar to gesture classification, our models for detecting erroneous gestures are trained using the Adam optimizer with step-decay and early stopping. We used low initial learning rates ranging from 0.0001 to 0.001 to help the stability of the optimization, given a small dataset. \par

\section{Experiments} 
We evaluated our monitoring system using trajectory data collected from the common surgical tasks of Block Transfer and Suturing performed in dry-lab settings on two different surgical platforms, the open-source Raven II surgical robot and the daVinci Research Kit (dVRK). The Raven II allowed us to simulate the impact of technical faults and attacks using software fault injection, while the surgical data collected from dVRK enabled studying the effect of human errors and evaluating the safety monitor using realistic surgical tasks. 

All experiments were conducted on an x86\_64 PC with an Intel Core i7 CPU @ 3.60GHz and 32GB RAM, running Linux Ubuntu 18.04 LTS, and an Nvidia 2080 Ti GPU, running CUDA 10.1. We used Keras \cite{chollet2015keras} API v.2.2.4 on top of TensorFlow \cite{abadi2016tensorflow} v.1.14.0 for training our models and Scikit-learn \cite{pedregosa2011scikit} v.0.21.3 for pre-processing and evaluation.

\subsection{daVinci Research Kit (dVRK)}
\textbf{JIGSAWS Dataset:}
\label{subsection: dvrk data}
For evaluating the performance of our solution on the dVRK, we considered the surgical task of Suturing. Since we did not have full access to the system, we used pre-collected trajectory data from the JIGSAWS dataset~\cite{gao2014jhu} and manually annotated the errors. The dataset contains synchronized kinematics and video data recorded at 30 Hz 
 from three surgical tasks (Suturing, Knot-tying and Needle-passing) that were performed by surgeons with varying skill levels in dry-lab settings on the dVRK platform. 
 The kinematics data comprises of 19 variables for each robot manipulator, including: Cartesian Position (3), Rotation Matrix (9), Grasper Angle (1), Linear (3) and Angular Velocity (3). 
We used the data from 39 demonstrations of the task of Suturing with the Leave-One-SuperTrial-Out (LOSO) setup of the JIGSAWS dataset for training and evaluating our monitoring pipeline. The LOSO setup meant that we trained on 4 super trials and held one super trial out for evaluation. 

\textbf{Erroneous Gesture Annotation:}
The gestures were already labeled as part of the JIGSAWS dataset \cite{gao2014jhu}, but we further classified them as safe or unsafe. 
We did so by manually annotating video data based on the rubric in Table \ref{tab:Rubric} and used it as ground truth for evaluating our classifiers which rely only on kinematics data. We labeled any given gesture as unsafe if any of the common errors specific to that gesture were observed in its corresponding video segment. Out of a total of 793 gestures, 144 were labeled as erroneous.  

\subsection{Raven II}

\begin{figure*} [t!]

    \captionsetup[subfigure]{justification=centering}
    \centering
     \begin{subfigure}[b]{0.25\textwidth}
    \centering
    \includegraphics[width=\linewidth ]{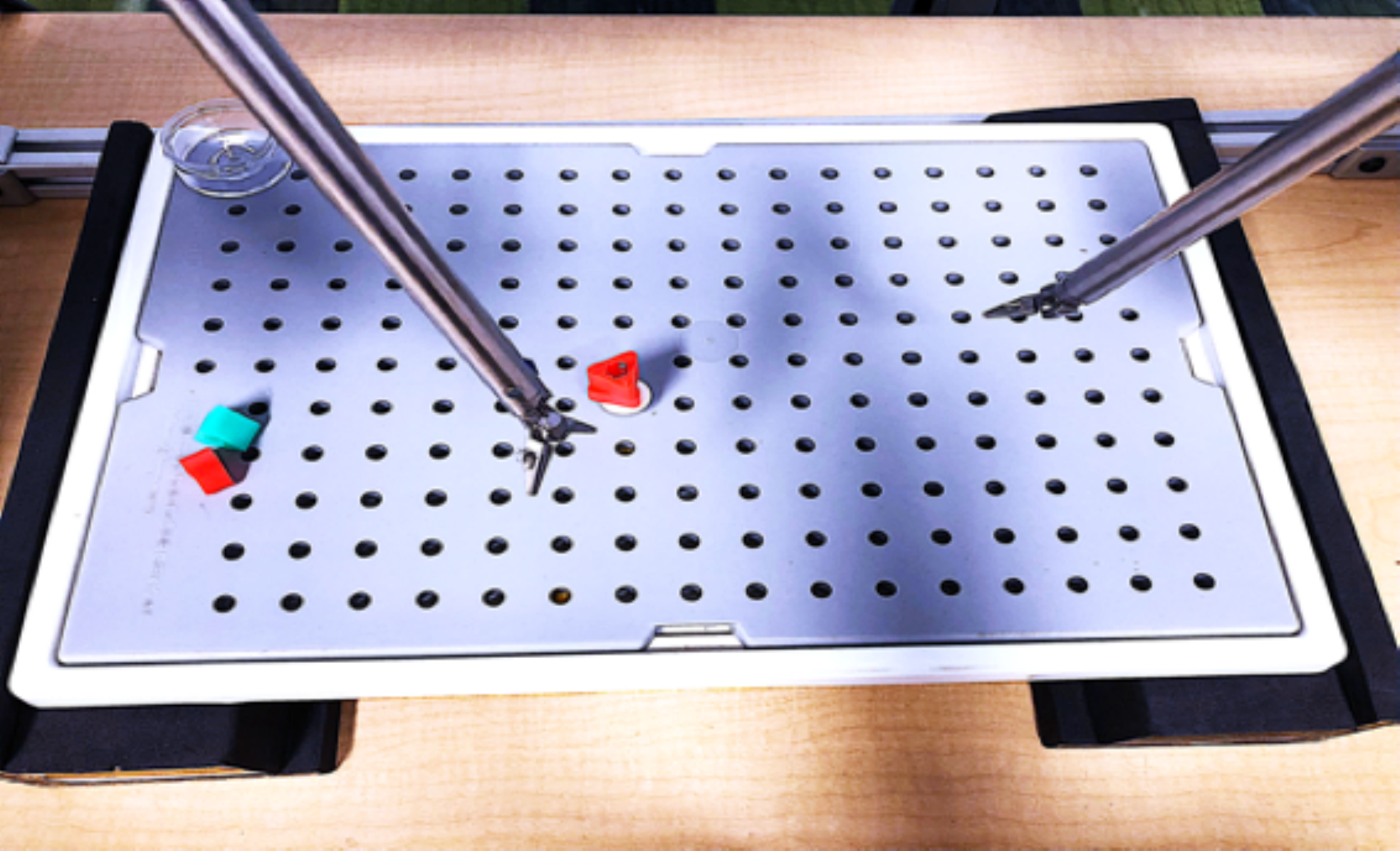}
    \caption{\centering Block Transfer in dry lab setting}
    \label{Fig:dry lab debridement}
    \end{subfigure}%
    \hfill
     \begin{subfigure}[b]{0.25\textwidth}
    \centering
    \includegraphics[width=\linewidth ]{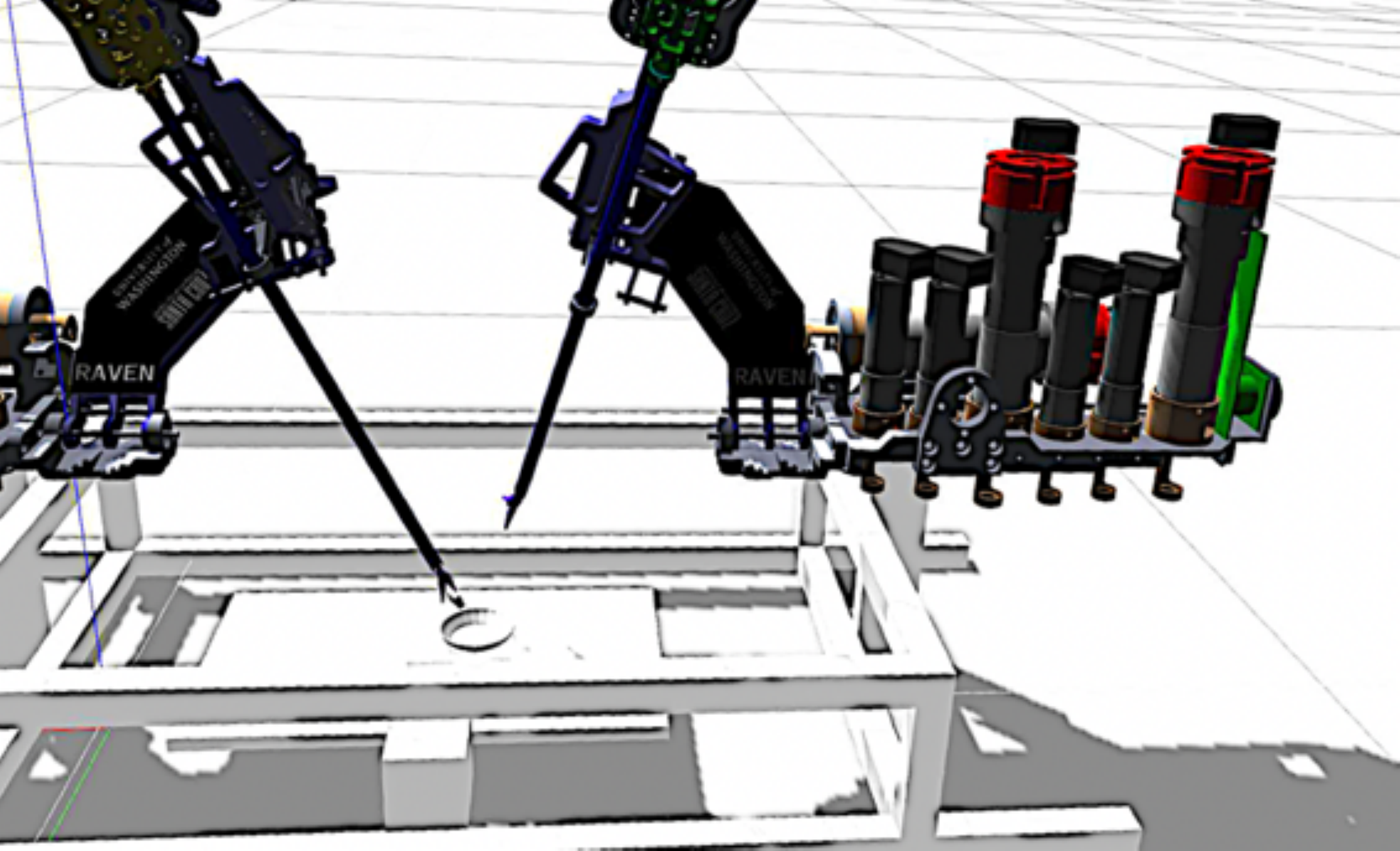}
    \caption{\centering Block Transfer in ROS Gazebo }
    \label{Fig:gazebo dry lab}
    \end{subfigure}%
    \hfill
     \begin{subfigure}[b]{0.25\textwidth}
    \centering
    \includegraphics[width=\linewidth ]{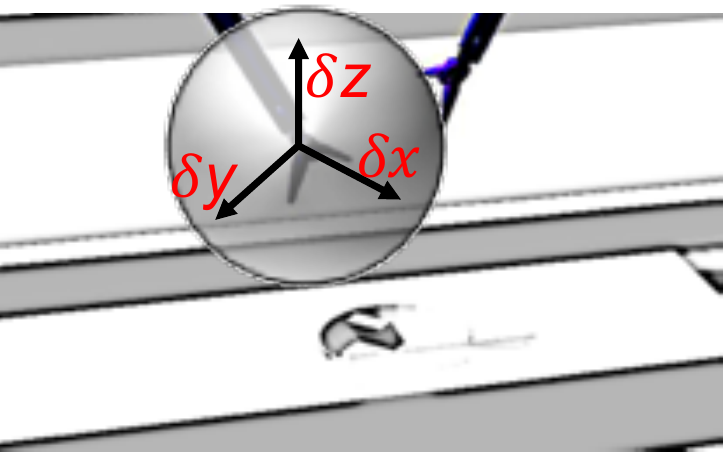}
    \caption{\centering Cartesian Position Faults}
    \label{Fig: gazebo cart fault}
    \end{subfigure}%
    \hfill
    \begin{subfigure}[b]{0.25\textwidth}
    \centering
    \includegraphics[width=\linewidth]{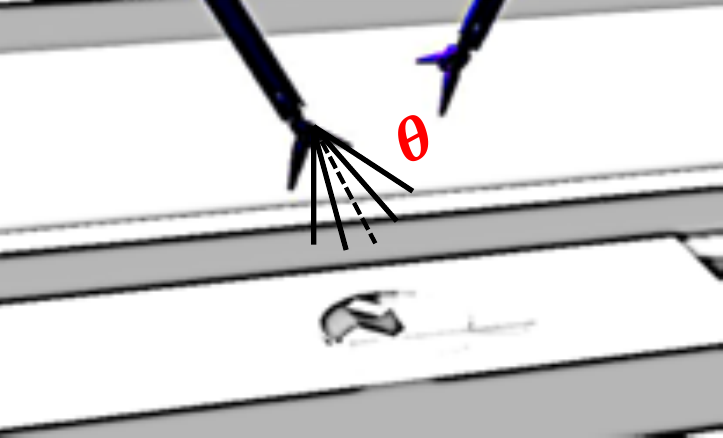}
    \caption{\centering Grasper Angle Faults}
    \label{Fig:gazebo grasper fault}
    \end{subfigure}
    \caption{Experimental setup for dry lab and virtual simulation of surgical tasks}
    \label{Fig:drylab lab gazebo }
   \vspace{-1em}
\end{figure*}

\textbf{ROS Gazebo Simulator:}
\label{subsubection: Gazebo Simulator}
For the Raven II, our experiments were conducted using a simulator that we developed based on ROS Gazebo 3D virtual environment, integrated with the RAVEN II control software and a fault injection tool that mimics the effect of technical faults and attacks in the robot control system. 
This simulator is available to the research community for experimental evaluation of safety monitoring solutions in robotic surgery\footnote{https://github.com/UVA-DSA/raven2\_sim/tree/gazebo\_sim}. 


We leveraged the physics engine of the Gazebo simulator for faithful representation of the dry-lab settings.  
Figures \ref{Fig:dry lab debridement}  and \ref{Fig:gazebo dry lab} show the dry-lab setup of the Block Transfer task in the Raven II workspace along with the corresponding simulation in the Gazebo 3D environment. All our experiments used the same setup with the left and right robot manipulators, grasper instruments, and the standard objects in the Block Transfer task, including a block and a receptacle where the block should be dropped. 

The input to the simulator can be surgeon's commands during tele-operation or output from motion planning algorithms in autonomous mode. The kinematics data from the simulator 
consisted of 277 features (including the 19 variables available from the JIGSAWS dataset), sampled at 1000 frames per second. The simulator also allows logging of the video data using a virtual camera. The video frames are logged at 30 frames per second, along with their timestamps to enable synchronization with kinematics data and to measure the times when faults and errors happen. We collected 20 fault-free demonstrations of the Block Transfer task performed by 2 different human subjects in the simulator, on which we carried out our fault injections. The dataset collected from the simulation experiments consisted of 115 fault-free and faulty demonstrations.


\textbf{Fault Injections:}
\label{subsection: fault injection}
We assume that accidental or malicious attacks and human errors can manifest as errors in the kinematic state variables in the inputs, outputs and the internal state of the robot control software and cause the common error types shown in Table~\ref{tab:Rubric}. As a result, our software fault injection tool directly perturbs the values of kinematic state variables to simulate such errors. Each injected fault is characterized by the name of the state variable ($V$) with value ($S$) that is targeted, along with the injected value ($S'$)  and the duration of the injection ($D$). 

Figure \ref{fig:blocktransferwithcontext} shows how the operational context for the Block Transfer task changes with respect to the kinematic state variables ($V$), which are the Grasper Angle and the Cartesian Position of the robot instrument end-effectors. The fault duration, $D$, is measured in milliseconds and could span across more than one gesture. 
We perturbed the values of Grasper Angle and the Cartesian Positions ($x, y, z$) in the collected fault-free trajectories 
and sent the faulty trajectory packets 
to the robot control software. This allowed us to repeat the same trajectory or to perturb only specific segments while the rest of the trajectory remained the same. 

To simulate the effect of attacks or human errors, we created deviations from the actual trajectory by slight increments or decrements in the values of Grasper Angle and the Cartesian Position variables. 
For Grasper Angle, we added a constant value of $\theta$ for the duration ($D$) until the target value ($S'$) was reached (see Figure \ref{Fig:gazebo grasper fault}). For Cartesian Position, we provided a target deviation ($\delta=d(S',S)$), which is the Euclidean distance between $S$ and $S'$ and a function of $x$, $y$ and $z$ values. 
We then enforced a uniform positive deviation in the three dimensions of $x$, $y$, and $z$ by injecting the value of $\delta_{x,y,z} = \delta/\sqrt[2]{3}$ to the three variables over the duration ($D$) (see Figure \ref{Fig: gazebo cart fault}).


\begin{table}[b]
    \centering
    \vspace{-2em}
    \resizebox{\linewidth}{!}{
    \begin{tabular}{|c|c|c|c|c|c|c|}
    \hline
          \multicolumn{5}{|c|}{Fault Types (Values, Durations, and Total Number) } & \multicolumn{2}{c|}{\# Errors (\%)} \\
         \hline
          Grasper & Duration & Cartesian Position & Duration & \# Fault  & Block-drop & Dropoff \\
          
          Angle (rad)& (\% Trajectory) &Deviation (mm) & (\% Trajectory)& Injections &  & Failure \\
         \hline
         \multirow{4}{*}{0.30-0.40} &  \multirow{2}{*}{0.55-0.70} & 3000-6000 &\multirow{2}{*}{0.50-0.60} & 16 & 0 (0\%) & \\
         & & 6000-65000 & & 8 &  1 (12.50\%) &\\ \cline{2-7}
         
         &\multirow{2}{*}{0.65-0.90} &  3000-6000 & \multirow{2}{*}{0.70-0.90} &16 & & 16 (100\%)\\
         & & 6000-65000 & & 16 & & 16 (100\%)\\
         
         \hline
         \multirow{4}{*}{0.50-0.60} & \multirow{2}{*}{0.55-0.70} & 3000-6000 &\multirow{2}{*}{0.50-0.60} & 16 & 0 (0\%) & \\
         & & 6000-65000 & & 8 &  1 (12.50\%) &\\\cline{2-7}
         
         &\multirow{2}{*}{0.65-0.90} &  3000-6000 & \multirow{2}{*}{0.70-0.90} &16 & & 16 (100\%)\\
         & & 6000-65000 & & 16 & & 15 (93.75\%)\\\cline{2-7}
         
         \hline
         \multirow{4}{*}{0.70-0.80} & \multirow{2}{*}{0.55-0.70} & 3000-6000 &\multirow{2}{*}{0.50-0.60} & 16 & 0 (0\%) & \\
         & & 6000-65000 & & 8 &  0 (0\%) &\\\cline{2-7}
         
         &\multirow{2}{*}{0.65-0.90} &  3000-6000 & \multirow{2}{*}{0.70-0.90} &16 & & 15 (93.75\%)\\
         & & 6000-65000 & & 16 & & 16 (100\%)\\\cline{2-7}
         \hline
          
         \multirow{4}{*}{0.90-1.00} & \multirow{2}{*}{0.55-0.70} & 3000-6000 &\multirow{2}{*}{0.50-0.60} & 58 & 28 (48.28\%) & \\
         & & 6000-65000 & & 50 & 33 (66\%) &\\\cline{2-7}
         
         &\multirow{2}{*}{0.65-0.90} &  3000-6000 & \multirow{2}{*}{0.70-0.90} &16 & 5 (62.50\%)& 6 (75\%)\\
         & & 6000-65000 & & 16 & 6 (75\%)& 6 (75\%)\\\cline{2-7}
         \hline
         
         \multirow{4}{*}{1.10-1.20} & \multirow{2}{*}{0.55-0.70} & 3000-6000 &\multirow{2}{*}{0.50-0.60} & 47 & 46 (95.78\%) & \\
         & & 6000-65000 & & 74 & 67 (86.49\%) &\\\cline{2-7}
         
         &\multirow{2}{*}{0.65-0.90} &  3000-6000 & \multirow{2}{*}{0.70-0.90} &16 & 12 (75.00\%) &\\
         & & 6000-65000 & & 16 & 12 (75.00\%) &\\\cline{2-7}
         \hline
         
         \multirow{4}{*}{1.30-1.40} & \multirow{2}{*}{0.55-0.70} & 3000-6000 &\multirow{2}{*}{0.50-0.60} & 41 & 40 (97.57\%) & \\
         & & 6000-65000 & & 61 & 58 (95.08\%) &\\\cline{2-7}
         
         &\multirow{2}{*}{0.65-0.90} &  3000-6000 & \multirow{2}{*}{0.70-0.90} &16 &14 (87.50\%)& \\
         & & 6000-65000 & & 16 & 14 (87.50\%)& \\\cline{2-7}
         \hline
         \multirow{4}{*}{1.50-1.60} & \multirow{2}{*}{0.55-0.70} & 3000-6000 &\multirow{2}{*}{0.50-0.60} &7 & 6 (85.71\%) & \\
         & & 6000-65000 & & 17 &  17 (100\%) &\\\cline{2-7}
         
         &\multirow{2}{*}{0.65-0.90} &  3000-6000 & \multirow{2}{*}{0.70-0.90} &16 & 16 (100\%)& \\
         & & 6000-65000 & & 16 & 16 (100\%)& \\
         \hline
         
         \hline
       
         \multicolumn{4}{|c|}{Total Fault Injections} & 651  & 392 & 106 \\

         \hline
    \end{tabular}
    }
    \caption{Fault injection experiments on the Raven II}
    \label{tab:Fault injection experiments}
    
\end{table}

Table \ref{tab:Fault injection experiments} shows the results of our fault injection experiments on the Gazebo simulator. In total, we conducted 651 fault injections in the task of Block Transfer, out of which, 498 resulted in errors, including 392 block-drop and 106 for dropoff failures. The state-space for injected values, $S'$, was 0.3 rad $\leq$$S'$$\leq$ 1.6 rad for Grasper Angle and 3000 mm $\leq$$S'$$\leq$ 65000 mm for Cartesian Position. We explored different combinations of perturbations of the targeted variables over different durations of the trajectory. 

The experiments showed that perturbing the Grasper Angle had a greater effect on causing errors compared to perturbing the Cartesian Position. For lower Grasper Angle values (0.3 rad $\leq$S'$\leq$ 0.8 rad), perturbation over different durations of the trajectory resulted in different failure modes. For fault injections with duration 0.65 $\leq$$D$$\leq$ 0.9 and targeted Grasper Angle 0.3 rad $\leq$$S'$$\leq$ 0.8 rad, the likelihood of a dropoff failure was high ($>$90\%) whereas for the same range (0.3 rad $\leq$$S'$$\leq$ 0.8 rad) but different duration (0.55 $\leq$$D$$\leq$ 0.7), the likelihood of any failure was significantly low. There were only 2 cases where the block was dropped at the wrong position due to high Cartesian deviation. When injecting higher values to the Grasper Angle (0.9 rad $\leq$$S'$$\leq$ 1.6 rad), we observed block-drops regardless of the duration of the perturbation, with higher values of $S'$ leading to higher percentage of failures. This suggests that for block-drop error to happen, the value of the Grasper Angle either needs to be higher than 0.8 rad, or the fault needs to be injected for a longer duration of time. For dropoff failure to happen, the Grasper Angle has to be below 1.0 rad, or the fault needs to be injected for a longer duration, possibly beyond G11, which is the gesture where the block should be dropped.

\begin{figure}[b!]
    \vspace{-1em}
    \centering
    \begin{subfigure}[b]{0.22\textwidth}
        \centering
        \includegraphics[width=\textwidth]{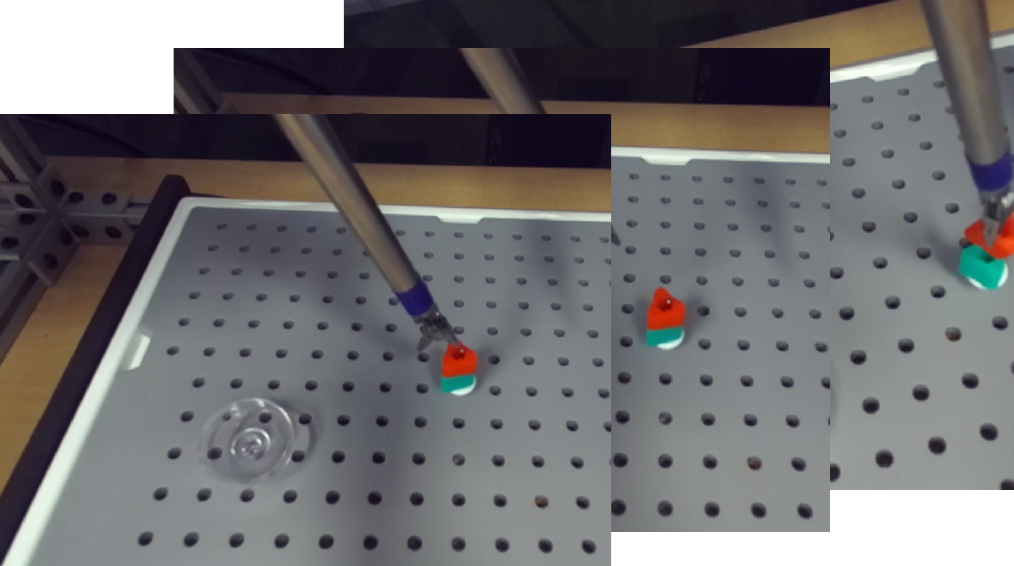}
        \caption[Video Frames]%
        {{\small Video Frames}}    
        \label{fig:input frames}
    \end{subfigure}
    \hfill
    \begin{subfigure}[b]{0.22\textwidth}  
        \centering 
        \includegraphics[width=\textwidth]{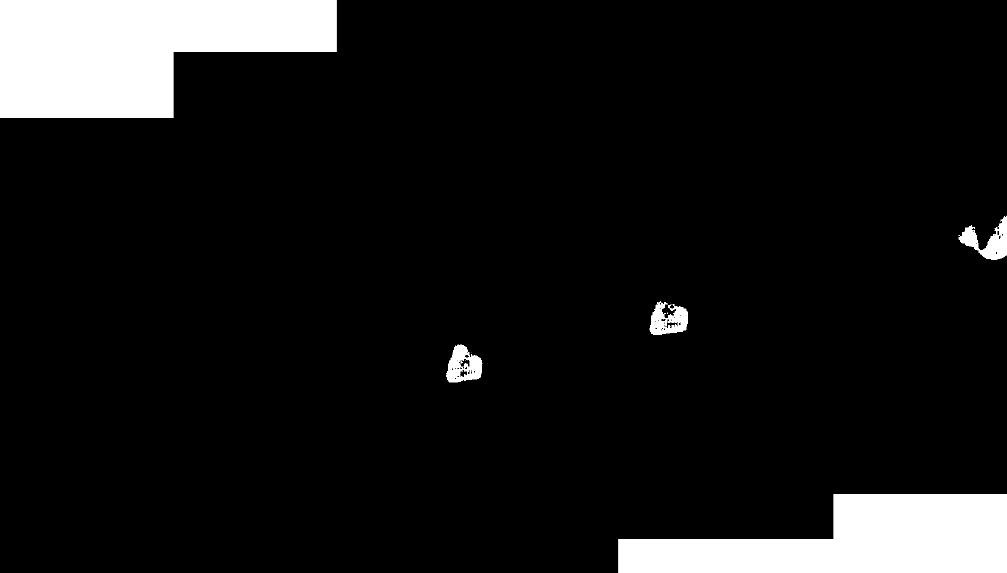}
        \caption[]%
        {{\small HSV Threshold of the block}}    
        \label{fig:hsv threshold}
    \end{subfigure}
    \begin{subfigure}[b]{0.22\textwidth}   
        \centering 
        \includegraphics[width=\textwidth]{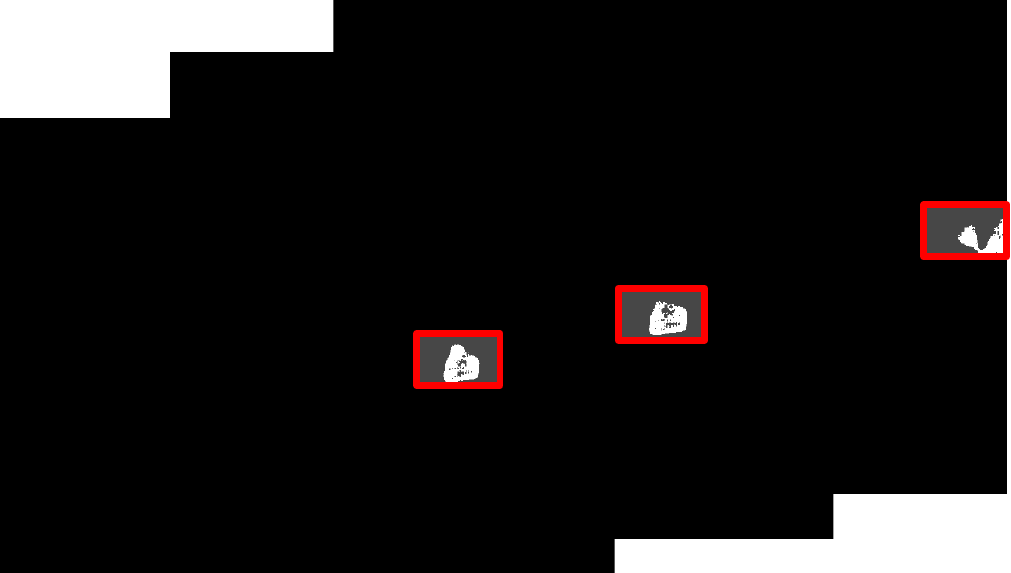}
        \caption[]%
        {{\small Contour Detection}}    
        \label{fig:contour detection}
    \end{subfigure}
    \quad
    \begin{subfigure}[b]{0.22\textwidth}   
        \centering 
        \includegraphics[width=1.2\textwidth]{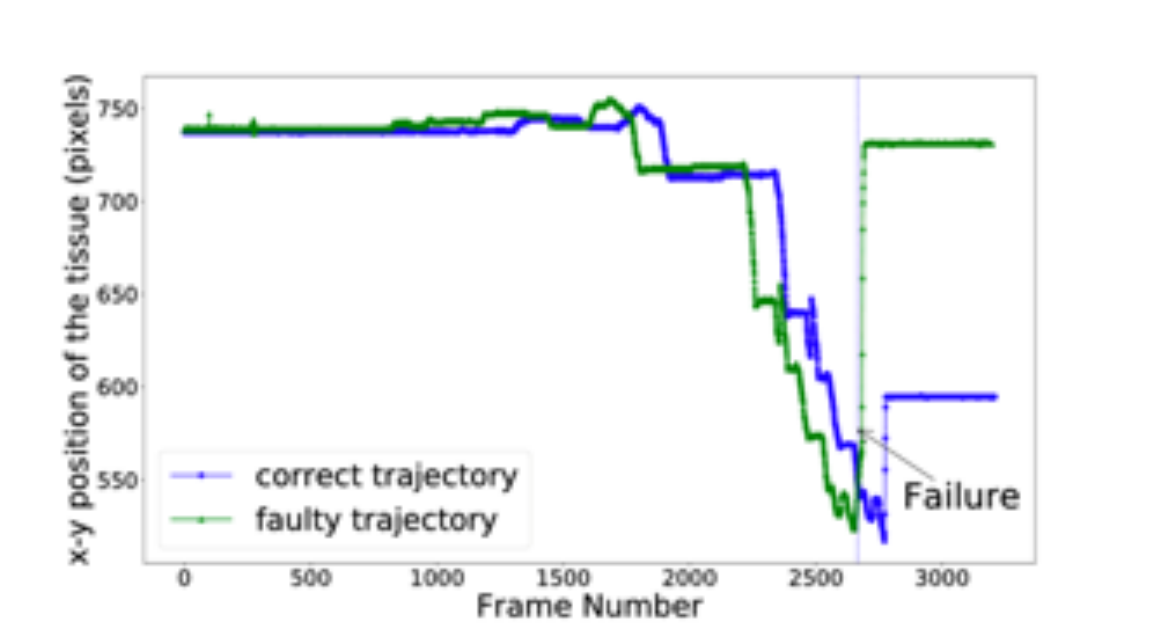}
        \caption[]%
        {{\small Comparison between traces, adopted from \cite{yasar2019context}}}    
        \label{fig:dtw trace}
    \end{subfigure}
    \caption[Failure detection using contour segmentation and DTW]
    {\small Failure detection using contour segmentation and DTW} 
    \label{fig:dtw failure}
\end{figure}

\textbf{Automated Labeling of Errors:}
We used computer vision approaches as an orthogonal method of detecting errors, as our fault injections were performed on the kinematics state variables. This, along with the knowledge of when a fault was injected, provided us the semantics to label a particular gesture as erroneous or non-erroneous. We adopted the marker-based (color and contour) detection approaches used in \cite{yasar2019context} here.
As the first step, we converted the logged video data to a sequence of frames (Figure \ref{fig:input frames}) with their corresponding timestamps.
For the case of detecting Block-drop, we used Structural Similarity Index (SSIM) \cite{wang2004image} on thresholded images (Figure \ref{fig:hsv threshold}) of the block to find the exact frame (and the timestamp) of when the failure happened. For the case of detecting Drop-off failure, we applied the same HSV threshold, followed by contour detection (Figure \ref{fig:contour detection}) to detect the contour of the block and track its centroid throughout the trajectory. We collected the trace of the centroid for the fault-free trajectories which we used as reference to compare against faulty trajectories. We used Dynamic Time Warping to compare the fault-free and faulty trajectory traces and checked for large deviations that indicate when the block should have been dropped, but it was not (Figure \ref{fig:dtw trace}). 

\textbf{Gesture Annotation:}
For annotating the data generated using the Gazebo simulator, we extended the data structure of the Raven II to include the current surgical gesture. This allows the human operator to record the surgical gesture as (s)he is simultaneously operating the robot, reducing the time and effort to look at videos and performing the annotations. For labeling the erroneous gestures, we recorded the time that we injected the fault to one of the kinematics state variables and the time that the fault led to any of the common errors in Table \ref{tab:Rubric} based on the video data and then mapped those times to the corresponding gestures. As a result, we were able to automate our gesture and erroneous gesture annotations for all experiments conducted in the Gazebo simulator. A total of 890 out of 4557 gestures were labeled as erroneous. \par

\subsection{Metrics}
We evaluated the individual components as well as the entire pipeline of our safety monitoring system in terms of \textbf{accuracy} and \textbf{timeliness} in identifying gestures and detecting errors using the metrics that are described next. 

\textbf{Individual Components:}
We trained individual components of the pipeline, namely the gesture classification and the erroneous gesture detection, separately. For the first part of the pipeline, our evaluation metrics were classification accuracy, for assessing model performance across different gesture classes, and jitter value, for identifying the timeliness of the classification. Jitter is calculated as the difference between the time our model detects a gesture and its actual occurrence, with \emph{positive} values indicating \emph{early detection}. 

Our evaluations of the second part of the pipeline were based on the standard metrics used for binary classification: True Positive Rate (TPR), True Negative Rate (TNR), Positive Predictive Value (PPV), and Negative Predictive Value (NPV), and the Area Under the ROC Curve (AUC) of the anomaly class. We reported the micro-averages for all the metrics unless stated otherwise. 

\textbf{Overall Pipeline:}
For evaluating the classification performance of the overall pipeline, we used the F1-score as well as the AUC of the negative class. In our case, it is imperative to not classify any erroneous gestures as non-erroneous (to not miss any anomalies), while keeping the False Positive Rate (FPR) low. The F1-score, which is the harmonic mean of precision and recall, is a good indicator of how the model performs in detecting or not missing erroneous gestures. At the same time, it only reports the performance of the model using one particular threshold. As F1-score is a point-based metric, we also used AUC of ROC curves, which reports the performance over different classification thresholds.  

Our metrics for assessing the timeliness of error detection were average computation time for the classifiers and \textit{reaction time}, defined as the time to react on the advent of an erroneous gesture and calculated as the difference between the actual time of error occurrence and the time it is detected: 
\begin{equation}
    reaction_t = actual_t - detected_t
    \label{equation: reaction_time}
    \vspace{-0.5em}
\end{equation}  

\begin{figure}[t]
    \centering
    
    \includegraphics[width=1\linewidth]{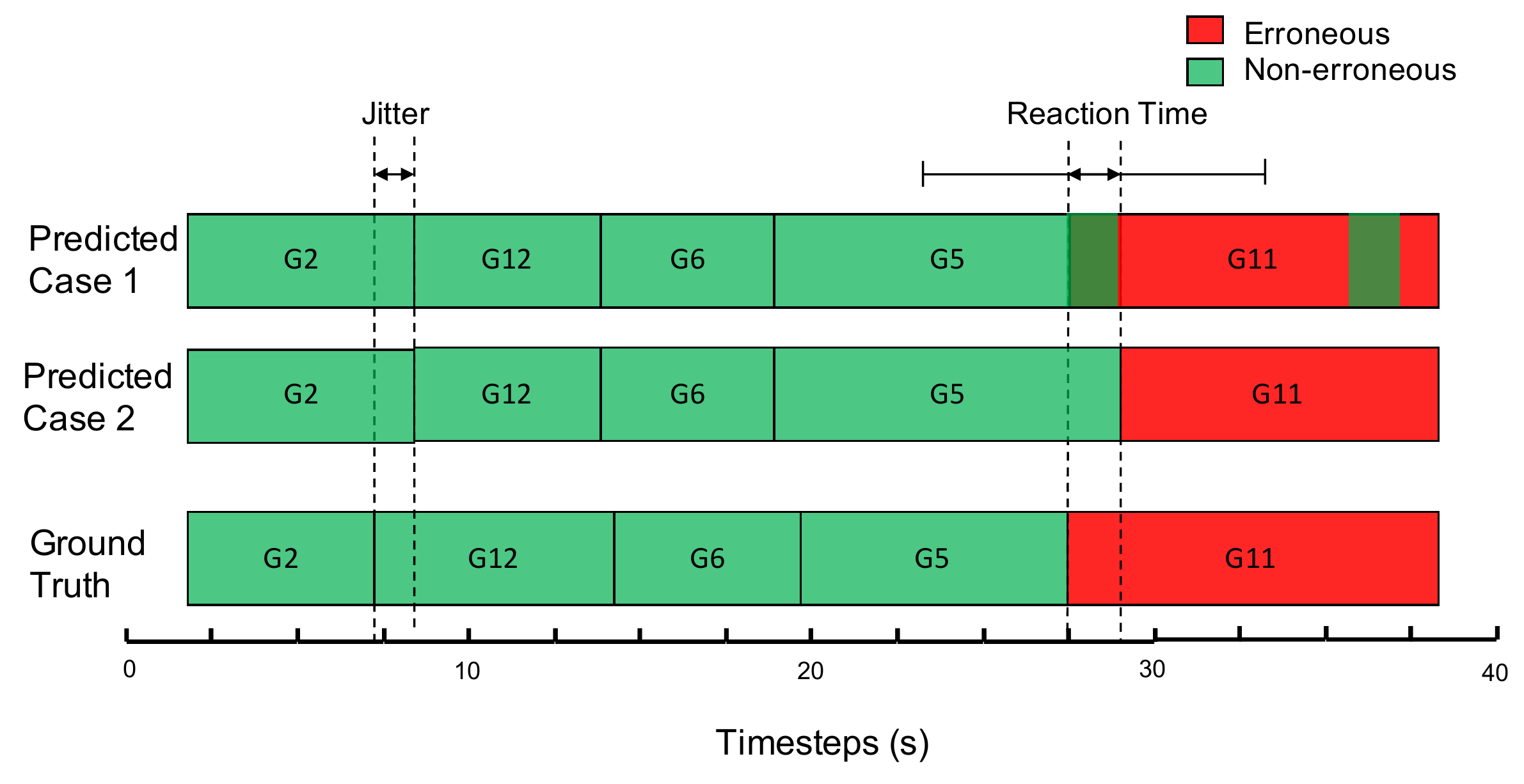}

    \caption{Example Timeline for Detecting Anomalies
    }
    \label{fig:timeline}
    \vspace{-2em}
\end{figure}

 The reaction time can be used as a measure of the time budget that we have for taking any corrective actions to prevent potential safety-critical events. A \emph{positive} value means that our model can predict an error \emph{before} its occurrence (early detection) whereas a negative value indicates the detection of error after it has already happened (detection delay). As shown in Case 1 in Figure \ref{fig:timeline}, our classifier predicts every kinematics sample as erroneous or non-erroneous. So there might be cases where different parts within the same gesture are classified as erroneous or non-erroneous. The reaction time is calculated based on the first time an erroneous sample is detected within a gesture. 

We also report the percentage of times that the erroneous gestures were detected before their actual occurrence (\% Early Detection in Table \ref{tab:whole_pipelineevaluation}). To calculate this, we divided the total number of times when the reaction time was positive by the total number of erroneous gesture occurrences.

\section{Results}
\label{section: results}

\subsection{Performance of Pipeline Components}

\textbf{Gesture Segmentation and Classification:} All our results are averaged across the 5 trials of LOSO setup. Table \ref{tab:gesture classification} shows the accuracy of our best performing model for all the  tasks in the JIGSAWS dataset compared to two state-of-the-art supervised learning models that only rely on kinematics data, \cite{lea2015improved} and \cite{sefati2015learning}. In addition, we also evaluated our model for the Block Transfer task on the Raven II. Our best performing model was a 2 layer stacked LSTM, with input time-step of 1, comprising of 512 and 96 LSTM units respectively, followed by a fully-connected layer with 64 units and a final softmax layer. For the tasks in the JIGSAWS dataset, the input to the model were all the 38 kinematics features from the robot manipulators. For the Block Transfer task on Raven II, we used the same LSTM architecture but the input to our model was the Cartesian Positions and Grasper Angles for each of the manipulators. \cite{lea2015improved} used a variation of the Skip-Chain Conditional Random Fields (SC-CRF) that can better capture transitions between gestures over longer periods of frames. \cite{sefati2015learning} introduced Shared Discriminative Sparse Dictionary Learning (SDSDL) that aims to jointly learn a common dictionary for all gestures in an unsupervised manner together with the parameters of a multi-class linear support vector machine (SVM). For Suturing, our gesture classifier achieved competitive average accuracy of 84.49\% on the test data. For Block Transfer, which has more training data and is a simpler task with no recurrence of gestures, our model achieved an accuracy of 95.16\%. 

Table \ref{tab:pipeline_effects} shows that for the Suturing task our model detected the gestures within a jitter value of 337 ms, performing  best for G2, G3, G4, and G6 with over 80\% accuracy and worst for G10. Our model was unable to detect G10 which is "Loosening more suture" as it does not occur frequently (see Fig. \ref{fig:cfg_suturing}), with only 1\% of transition probability from G6 and 13\% transition probability from G4. In addition, as seen in Table \ref{tab:Rubric}, there were no common errors in G10.

\begin{table}[b!]
    \vspace{-1em}
    \centering
    \resizebox{\linewidth}{!}{
    \begin{tabular}{|c|c|c|c|c|}
    \hline
    Method & Suturing & Knot Tying & Needle Passing & Block Transfer \\
    \hline
         \textbf{This work} &  84.49 \% & 81.69 \% & 69.34 \% & 95.16 \%\\
     \hline
         SC-CRF \cite{lea2015improved} & 85.24 \% & 80.64 \% & 77.47 \% & N/A\\
    \hline
         SDSDL \cite{sefati2015learning}  & \textbf{86.32 \%} & 82.54 \%& 74.88 \% & N/A\\
    \hline 
    Training size &102,698 & 44,512 & 66,914 & 4,197,988\\ 
    \hline
    Number of Trajectories &39 &28 &36& 115 \\
    \hline
    \end{tabular}
    }
    \caption{Gesture classification accuracy in LOSO setup}
    \vspace{-1em}
    \label{tab:gesture classification}
\end{table}

\begin{table}[t!]
    \begin{threeparttable}
    
    \centering
    \resizebox{\linewidth}{!}{
    \begin{tabular}{|p{0.1\linewidth}|p{0.08\linewidth}|p{0.1\linewidth}|p{0.09\linewidth}|p{0.06\linewidth}|p{0.04\linewidth}|p{0.04\linewidth}|p{0.04\linewidth}|p{0.04\linewidth}|}
    
        \hline
        
         Setup & Model & Layers & Features & Lr \tnote{a} & TPR & TNR & PPV & NPV \\
         \hline
         gesture specific&LSTM & 512,128, 64,16\tnote{*} & All & 1e-4 & 0.75 &0.72 & 0.67& 0.80 \\
         \hline
         gesture specific& LSTM & 128,32, 16,16\tnote{*}& C,R,G & 1e-4 &\textbf{0.76} & 0.72 & 0.67&  \textbf{0.81}  \\
         \hline
         gesture specific & Conv &512,128, 32,16\tnote{*}& C,R,G & 1e-4 & 0.76& 0.73 & 0.68 & 0.80    \\
         \hline
         gesture specific& Conv & 512,128, 32,16\tnote{*}& All & 1e-4 & \textbf{0.76} & \textbf{0.73} & \textbf{0.69}& 0.80    \\
         \hline
         non-gesture specific&LSTM & 512,128, 64,16\tnote{*}& All & 1e-4 & 0.73& 0.71  & 0.66 & 0.77    \\
         \hline
    \end{tabular}}
    \caption{Overall performance of the erroneous gesture classification step for Suturing on the dVRK using different setups, for input time-window=5, stride=1} 
    \label{tab:evaluation_hyper-parameters_dvrk}
    \begin{tablenotes}\footnotesize
    \item [a] Initial Learning Rate, * Fully-Connected Layer
    \end{tablenotes}
    \end{threeparttable}
    \vspace{-2em}
\end{table}

\begin{table}[b]
    \vspace{-1em}
    \begin{threeparttable}
    
    \centering
    \resizebox{\linewidth}{!}{
    \begin{tabular}{|p{0.1\linewidth}|p{0.08\linewidth}|p{0.1\linewidth}|p{0.09\linewidth}|p{0.06\linewidth}|p{0.04\linewidth}|p{0.04\linewidth}|p{0.04\linewidth}|p{0.04\linewidth}|}
    
        \hline
        
         Setup & Model & Layers & Features & Lr \tnote{a} & TPR & TNR & PPV & NPV \\
         \hline
         gesture specific& Conv & 256,128, 64,16\tnote{*}& C,G & 1e-4 &\textbf{0.62} & \textbf{0.87} & \textbf{0.65} &  0.86  \\
         \hline
         gesture specific& LSTM & 64,32, 64,16\tnote{*} & C,G & 1e-4 & 0.62& 0.85 & 0.57 & \textbf{0.89}\\
         \hline
         non-gesture specific& Conv & 256,128, 64,16\tnote{*} & C,G & 1e-4 & 0.59& 0.85 & 0.58 &0.85    \\ %
         \hline
    \end{tabular}}
    \caption{Overall performance of the erroneous gesture classification step for Block Transfer on the Raven II using different setups, for input time-window=10, stride=1} 
    \label{tab:evaluation_hyper-parameters_ravenII}
    \begin{tablenotes}\footnotesize
    \item [a] Initial Learning Rate, * Fully-Connected Layer
    \end{tablenotes}
    \end{threeparttable}
    \vspace{-1em}
\end{table}

\par
\textbf{Erroneous Gesture Detection:}
\label{section: performance anomaly detection}
We trained our erroneous gesture detection system on individual gestures, assuming perfect gesture boundaries. This allowed us to independently evaluate the performance of this module and evaluate different architectures and models that are suited for time-series classification, including LSTM networks and 1D CNNs. We also experimented with different supervised learning architectures, from kernel-based models such as SVM to ensemble techniques such as Random Forest, but here only report results for LSTM networks and 1D-CNNs for their superior performance over other architectures. We further experimented with different subsets of kinematics features, while using the set of all the features as our baseline. Our experiments specifically involved using different combinations of Cartesian Position (C), Rotation matrix (R), Grasper Angle (G) and Joint Angle (J) variables. 

Tables \ref{tab:evaluation_hyper-parameters_dvrk} and \ref{tab:evaluation_hyper-parameters_ravenII} show the best performing models for each setup for Suturing and Block Transfer tasks, respectively. We overall observed that being gesture specific led to better accuracy (higher TPR, TNR, PPV, NPV), even with smaller datasets and that 1D-CNNs performed better than LSTM networks for binary classification of gestures for both the tasks. Training the models using specific features (Cartesian, Rotation and Grasper Angle) led to similar or better performance compared to training with all the features. 
In both cases, the best performing model had higher TPR and TNR while achieving competitive NPV and PPV. This suggests that the models can identify the unsafe gestures with good accuracy while not providing too many false alerts. 

Table \ref{tab:auc for each gesture} shows the average AUCs achieved for each gesture class using the best performing 1D-CNN model. 
Our model performed best on gestures G6 and G4 for Suturing. The common error for both was when the "Robot end-effector is out of sight", which occurred frequently in the demonstrations and often among surgeons with less expertise. For Block Transfer, G6 had the highest accuracy, although the common gesture-specific error in this case is "Unintentional needle/object drop". We also measured the average reaction time for detecting erroneous gestures for each gesture, as shown in Table \ref{tab:pipeline_effects}. In our setup, the best value would be 0, which is when the detection of erroneous gesture coincides with the start of the gesture, due to the design of our pipeline where we first detect the gesture and then the gesture-specific anomaly, if any. Since we had erroneous and non-erroneous gestures, we also calculated the average jitter for erroneous gestures, to see their difference when compared to the overall average jitter and their effect on the reaction time. For Suturing, our model performed best for gesture G4, with an average reaction time of -0.01 frames (- 0.34 ms), as well as competitive average jitter for erroneous gestures of -84 ms, followed by G1 which had an average reaction time of -6.0 frames (-167 ms). Gestures G10 and G11 had no common errors and hence no reaction times. 
When looking across all gestures, we noticed our model performed best for gestures which are commonly occurring in the Suturing and Block Transfer tasks and also have higher number of errors. Improvement over less common gestures, with sparse errors will be the focus of future work. 

\begin{table}[t]
    \centering
    \resizebox{\linewidth}{!}{
    \begin{tabular}{|c|c|c|c|c|c|}
    \hline
        Gesture & Train Size & \% Errors & Test Size  & \% Errors & AUC\\
    \hline
        G1 & 1432 & 29 & 358 & 28 &  0.60\\
    \hline
        G2 & 13728 & 25 & 3432 & 24 &  0.50\\
    \hline
        G3 & 34921 & 41 & 8731 & 40 & 0.70\\
    \hline
        G4 & 13339 & 77 & 2601 & 79 & \textbf{0.93}\\
    \hline
        G5 & 2717 & 5 & 680 & 4 & 0.61\\
    \hline
        G6 & 18923 & 74 & 4731 & 74 & \textbf{0.93} \\
    \hline
        G8 & 8413 & 45 & 2104 & 29 & 0.81\\
    \hline 
        G9 & 1769 & 59 & 443 & 56 &  0.61\\
    \hline
    \hline
   
    G5 & 681976 & 24 & 151038 & 19 & 0.72\\
    \hline
     G6 & 394077 & 25 & 88748 & 21 & \textbf{0.75}\\
    \hline
    G11 & 241067 & 53 & 53969 & 41 & 0.66\\
    \hline
    \end{tabular}}
    \caption{Performance of the  erroneous gesture classifiers}
    \label{tab:auc for each gesture}
    \vspace{-2em}
\end{table}

\vspace{-0.5em}
\subsection{Overall Performance of Safety Monitoring Pipeline}
\label{section:evaluation of whole pipeline}
We evaluated the overall performance of our Safety Monitoring pipeline for two different setups of gesture-specific and non-gesture-specific. Although we used offline data for our analysis, our system can perform the classification in real-time.

\textbf{Non-Context-Specific Safety Monitoring:}
As a baseline, we trained a classifier with no explicit notion of context in terms of training labels, by feeding it only the kinematics data and the corresponding safe/unsafe labels. Due to the the ability of LSTM networks to recognize varying spatio-temporal patterns coupled with larger data sizes compared to gesture-specific classifiers, the classifier demonstrated some generalization and attained competitive performance (see Table \ref{tab:whole_pipelineevaluation}). An average F1-score of 0.72 and AUC of 0.71, reaction time of +6.62 frames or 221 ms, and computation time of 1.9 ms was achieved for Suturing. For Block Transfer, the classifier achieved an average F1-score and AUC of 0.73 and 0.74, respectively. The reaction time was -15.2 frames or -457 ms. 

\textbf{Context-Specific Safety Monitoring:}
\label{section: performance no-context}
In this setup, the input kinematics samples were first passed to the gesture classifier step. Having detected the gestures, their corresponding kinematics samples were sent to a separate gesture-specific classifier to identify their safety properties. 
As seen in Table \ref{tab:whole_pipelineevaluation}, for Suturing the average F1-score and AUC were 0.76 and 0.81, respectively, which is an improvement over the results obtained with non-context-specific setup. The average reaction time was -1.7 frames (-57 ms) and average computation time was 2.1 ms. For Block Transfer, the trend in accuracy is similar, with the gesture-specific setup achieving a higher average F1-score of 0.88 versus 0.73 and a higher AUC of 0.86 versus 0.74. The higher accuracy, as reflected by F1-score and AUC, provides more evidence (in addition to distribution analysis in Section \ref{subsection: distribution analysis}) to support our hypothesis about the context-specificity of errors. 

\begin{table}[b!]
    \centering
    \resizebox{\linewidth}{!}{%
    \begin{tabular}{|p{0.35\linewidth}|p{0.07\linewidth}|p{0.09\linewidth}|p{0.1\linewidth}|p{0.14\linewidth}|p{0.12\linewidth}|}
        \hline
         Setup & Avg. AUC & Avg. F1 &Avg. React Time (ms)  & Early Detection (\%) & Avg. Compute Time (ms) \\
         \hline
         Gesture-specific with perfect gesture boundaries for Suturing & \textbf{0.83 $\pm{.14}$} & \textbf{0.79} $\pm{0.13} $& +53 $\pm{797}$ & 38.89 \% & N/A\\
         \hline
          \textbf{Gesture-specific with gesture classifier} for Suturing& 0.81 $\pm{.14}$ & 0.76 $\pm(0.13)$ &-57 $\pm{1030}$ & \textbf{43.75 \%} & 2.1 \\
         \hline
         Non-gesture-specific classifier for Suturing & 0.71 $\pm{.16}$ & 0.72 $\pm{0.12}$ & \textbf{+221} $\pm{1047}$ &  34.53 \% & 1.9 \\
         \hline
         \hline
         \textbf{Gesture-specific with gesture classifier} for Block Transfer & \textbf{0.86} $\pm{.15}$ & \textbf{0.88} $\pm{.14}$  &-1693 $\pm{5670}$ & 28.26 \% & 3.2\\
         \hline
         
         Non-gesture-specific classifier for Block Transfer & 0.74 $\pm{.18}$ & 0.73 $\pm{0.17}$ & \textbf{-457} $\pm{4520}$ &  \textbf{38.78} \% &1.5\\
         \hline
    \end{tabular}}
    \vspace{-0.5em} 
    \caption{Evaluation of the overall pipeline with ground-truth vs. predicted gestures, compared to a non-gesture-specific approach}
    
    \label{tab:whole_pipelineevaluation}
\end{table}

The gesture-specific models had comparatively worse reaction and computation times than the non-gesture specific pipeline due to the latency introduced for identifying the context before detecting gesture-specific anomalies. Figure \ref{fig:timeline} (Case 2) provides examples of how a negative jitter associated with the detection of the gesture can result in negative reaction times. However, for Block Transfer, we were only late by -50.8 frames or 1693 ms and for Suturing, by 220 ms, while having high accuracy. 

\begin{figure}[b]
    \vspace{-2em}
    \centering
      \includegraphics[trim=0 0 0 30, clip ,width=0.85\linewidth]{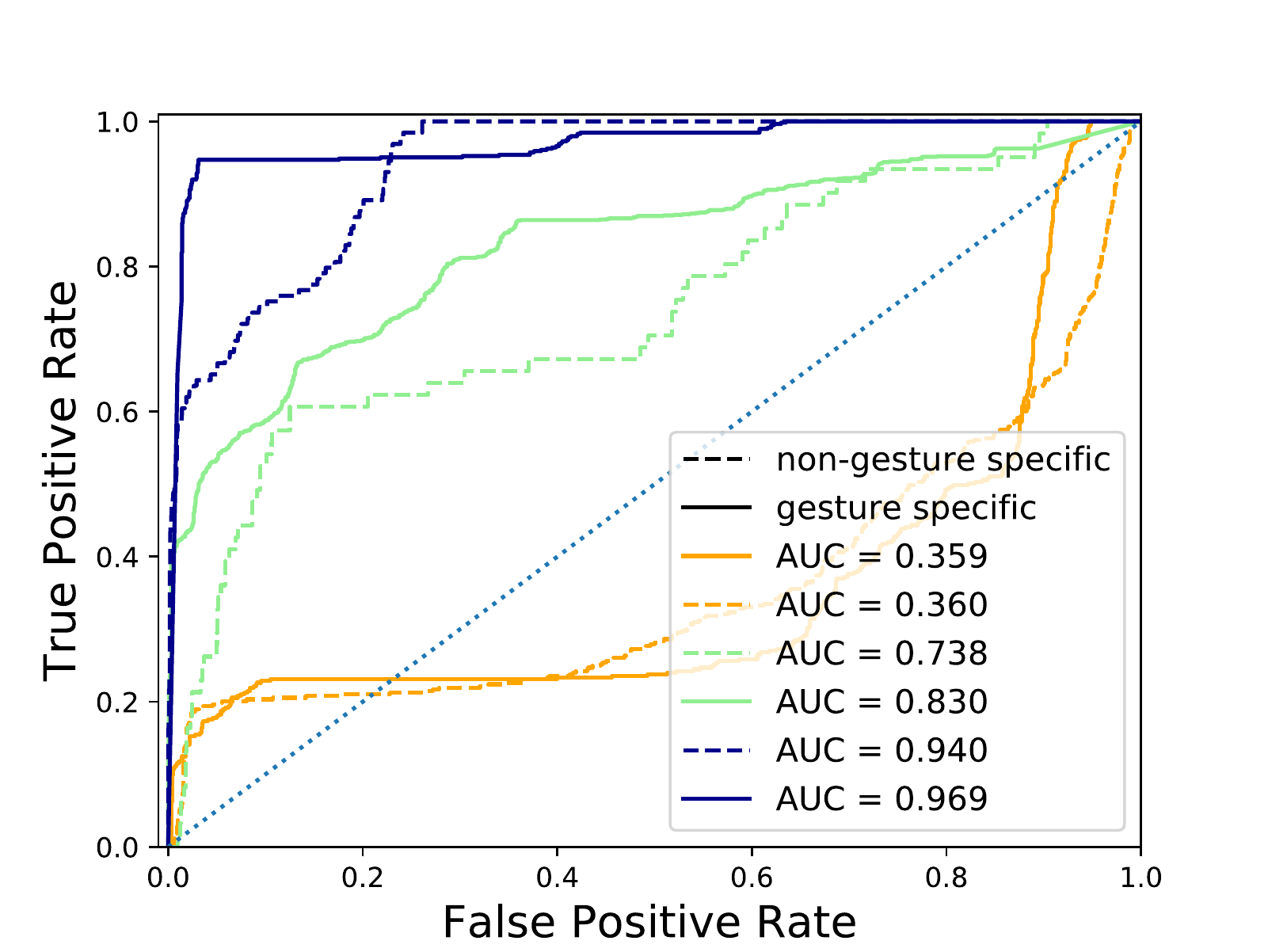}
    \vspace{-0.5em}
    \caption{Best, median and worst ROC curves for the whole pipeline in non-context-specific (baseline) and context-specific setups}
    \label{fig:ROC_MinMaxMedian}
    \vspace{-2em}
\end{figure}

Figure \ref{fig:ROC_MinMaxMedian} compares the worst, best and median performance of the context-specific and non-context-specific setups across different demonstrations, with the context-specific pipeline having an overall better performance. To get an empirical upper bound for the overall performance of the pipeline, we evaluated our entire pipeline assuming perfect gesture boundaries. As shown in Table \ref{tab:whole_pipelineevaluation}, when using perfect gesture boundaries, the average AUC improved from 0.81 to 0.83 and reaction time improved from -57 ms to 53 ms.


\begin{table*}[]
    \centering
    \begin{tabular}{|c|c|c|c|c|c|c|c|}
    \hline
        \multirow{2}{*}{Gesture}
         & \multicolumn{2}{c|}{Perfect Boundaries} & \multicolumn{5}{c|}{Gesture Specific Pipeline} \\
         \cline{2-8}
         & Reaction Time  & F1 score & Average Jitter & Gesture Detection  & Average Jitter for  & Reaction  & F1 score for  \\
         &(ms)  & erroneous gestures & (ms)&  Accuracy &erroneous gestures (ms) &  Time (ms) &  erroneous gestures \\
         \hline
         G1 & -2050 & 0.69 & 147 & 45.5 & -2317 & -167 & 0.63\\
         \hline
         G2 & -189 & 0.36 &-110 & 81.1 & -95 & -703 & 0.33\\
         \hline
         G3 &  -1810 & 0.54 &180 & \textbf{90.4} & -370 & -2574 & 0.45\\
         \hline
         G4 &  \textbf{0} & \textbf{0.94} & -154 & 86.7 &-84 & \textbf{-0.34} & \textbf{0.90}\\
         \hline
         G5 &  0 & 0  & -130 & 71.2 & \textbf{0} & -2367 & 0.09\\
         \hline         
         G6 &  -146 & 0.94 & -124 & 87.1 & -136 & -235 & 0.90\\
         \hline         
         G8 &  -970 & 0.66 & -337 & 67.4 & -1842 & -610 & 0.60\\
         \hline         
         G9 &  -377 & 0.76 & \textbf{46} & 63.8 & -474 & -767 & 0.60\\
         \hline         
         G10 &  N/A & N/A & N/A & N/A &N/A &N/A & N/A\\
         \hline         
         G11 &  N/A & N/A & 297 & 76.8 & N/A & N/A & N/A  \\
         \hline
         \hline
         G2 &  N/A& N/A & 397 & \textbf{96.2} & N/A &N/A & N/A\\
         \hline
         G5 & -708  & 0.75 & 440 &  95.1& -228 & -2127 & 0.58 \\
         \hline
         G6 & -1562 & 0.80 & 38  & 96.1&-137 &-2283 &0.70 \\
         \hline
         G11 & \textbf{-307} & \textbf{0.94} & -620 &80.1 & \textbf{-85} &\textbf{-667} & \textbf{0.73}\\
         \hline         
         G12 & N/A &  N/A& \textbf{6} &  92.6&N/A &N/A & N/A \\
         \hline
    \end{tabular}
    \caption{Effect of the pipeline components on the accuracy}
    \label{tab:pipeline_effects}
    \vspace{-2em}
\end{table*}

{}

\section{Discussion}
\label{section: discussion}
Our results provide encouraging evidence for the possibility of accurate and timely detection and possible preemption of erroneous gestures. Our experiments provided us with a number of key insights: 

\textbf{Being context-specific results in more accurate detection of erroneous gestures but worse reaction times}. Table (\ref{tab:whole_pipelineevaluation}) shows that there is an improvement of 14.1\% and 16.2\% of AUC over non-context specific detection, for Suturing and Block Transfer tasks, respectively. Having the notion of context reduces the search-space for erroneous gestures, hence allowing our models to have better TPR/TNR. However, we note that finding the best trade-off between TPR/FPR from the ROC while considering the implications for surgical safety is non-trivial and requires data from real surgeries, including adverse events and close feedback from surgeons. On the other hand, the gesture-specific models result in negative reaction times (later detection of anomalies) and higher computation times due to the latency introduced for identifying the context. However, the average reaction times are still within the 1-1.5 seconds time frame. Thus, there are still opportunities for issuing timely alerts or corrective actions (when the error periods are longer) and for the acceleration of context inference stage and improving the reaction times.

\textbf{Error detection with no notion of context achieves competitive performance.} With an AUC of 0.71 and 0.74 for Suturing and Block Transfer tasks, the non-gesture-specific models can be considered as a good baseline. However, their high accuracy is partly due to using larger training size data (samples from all the gesture classes) and learning from similar error patterns in some of the gesture classes. 
     
\textbf{Gesture classification performance does not proportionally impact the overall error detection performance.} In other words, some errors can still be detected even if the gestures are mis-classified. This is because some gestures have very similar error patterns or common failure modes. For example, for Suturing, the gestures G4 and G6 both have the same failure mode where "the needle holder is not in view at all times". 
     
\textbf{Having perfect gesture boundaries leads to improved AUC and reaction time}. When we look at the effect of the gesture classifier on the erroneous gesture detection (Table \ref{tab:pipeline_effects}), we see that for all the gestures, having perfect classification boundaries would have resulted in better reaction times and F1 scores for erroneous gestures. This suggests possible scope of improvement in the direction of gesture classification, while also suggesting that possibly predicting the gesture boundary ahead of time could result in better reaction time. For Suturing in particular, the gestures with the highest F1 score for the gesture specific pipeline were G4 and G6, which also had high gesture detection accuracy.
    
\textbf{Higher F1 score for detecting erroneous gestures has the highest impact on the reaction time.} As seen in Table \ref{tab:pipeline_effects}, misclassifying gestures or negative jitter values have less impact on the reaction time. On the other hand, the best reaction time is -0.34 ms for G4, which also had the highest F1 score for detecting erroneous gestures. 
    

\textbf{1-D CNN performs better than LSTM models for detecting erroneous gestures}. Firstly, we are only classifying kinematics samples within a gesture to safe or unsafe, instead of across the entire trajectory, meaning that there is no long/short-term dependency over the class. Secondly, 1D-CNNs benefit from the feature extraction of the Convolutional layers to learn a good mapping between the gesture-specific patterns and the binary labels. Combining Convolutional layers with LSTM units would greatly increase the computational cost of the pipeline and potentially the timeliness of the monitor, thus, it was not considered here.

\section{Related work}
\label{section: related work}
\par
\textbf{Safety and Security in Medical Robotics}:  
Safety is widely recognized as a crucial system property in medical robotics.  
Previous work \cite{jung2014safety} 
introduced a conceptual framework that can capture both the design-time and run-time characteristics of safety features of medical robotic systems in a systematic and structured manner. In \cite{alemzadeh2014systems}, a systems-theoretic hazard analysis technique (STPA) was used to identify the potential safety hazard scenarios and their contributing causes in the RAVEN II and the corresponding real adverse events reported for daVinci surgical robot \cite{alemzadeh2016adverse}. \cite{alemzadeh2016targeted} proposed an anomaly detection technique based on real-time simulation of surgical robot dynamic behavior and preemptive detection of safety hazards such as abrupt jumps of end-effectors. In \cite{yasar2019context}, we presented a monitoring system for real-time identification of subtasks using unsupervised techniques and detecting errors based on subtask-specific safety constraints learned from a very small set of fault-free demonstrations. \cite{he2019enabling} presented a system to predict unsafe manipulation in robot-assisted retinal surgery by measuring small scleral forces and predicting force safety status. 

Coble et al. proposed using remote software attestation for verification of potentially compromised surgical robot control software in unattended environments such as the battlefield~\cite{coble2010secure}. Other works have focused on improving the security of surgical robots by introducing new networking protocols 
such as Secure and Statistically Reliable UDP (SSR-UDP)~\cite{tozal2011secure} and Secure ITP~\cite{lee2012cyberphysical} that aim at increasing reliability and confidentiality of surgeon's commands.
 
This paper has the similar 
goal of early detection and mitigation of safety-critical events as \cite{alemzadeh2016targeted, yasar2019context}, but it uses \emph{supervised deep learning} methods for \emph{online} and \emph{gesture-specific} safety monitoring. It can be used in conjunction with the mechanisms proposed by the previous works to improve resilience of surgical robots against both errors and attacks.


\textbf{Surgical Workflow Analysis:} 
Automatic analysis of surgical workflow for surgeon skill evaluation and surgical outcome prediction has been the subject of many previous works. 
In \cite{rosen2006generalized}, authors modeled minimally invasive procedures as stochastic processes using Markov chains and used kinematics data along with dynamics of surgical tools for decomposing complex surgical tasks. \cite{lin2006towards} presented a feature collection, processing and classification pipeline for automatic detection and segmentation of surgical gestures (surgemes) in dry-lab settings. \cite{dipietro2016recognizing} showed that Recurrent Neural Networks can be used for the task of gesture recognition, while maintaining smooth boundaries over time. In \cite{zia2018surgical}, authors proposed RP-Net, a modified
version of InceptionV3 model \cite{szegedy2016rethinking}, for automatic surgical activity recognition during robot-assisted radical prostatectomy (RARP) procedures.
\cite{katic2013context} combined formal knowledge, represented by an ontology, and experience-based knowledge, represented by training samples, to recognize current phase of a surgery for context-aware information filtering. In this work, we focus on modeling the surgical context similar to the Markov chain models presented in \cite{rosen2006generalized} and on identifying the surgical gestures based on time-series data similar to \cite{dipietro2016recognizing}. However, our main goal is to detect the \emph{erroneous gestures}. 

\textbf{Context-Aware Monitoring:}
Context-aware anomaly detection has been the focus of many recent works on safety-critical systems. For example, in \cite{yuan2014context} a context-aware reasoning framework with sensor data fusion and anomaly detection mechanisms was developed to support personalized healthcare services at home for the elderly.
\cite{zoppi2016context} showed that using the notion of context and incorporating usual behavior of services leads to improved detection accuracy over traditional detection mechanisms for critical service oriented architectures. \cite{duessel2017detecting} provided a framework for context-based detection of network intrusions by incorporating protocol context and byte sequences. Our work shares similarities with the aforementioned by incorporating context for improving anomaly detection, but it relies on deep learning for \emph{real-time context-inference and anomaly detection} in robotic surgery.  

\section{Threats to Validity}
Our solution relies on the accuracy and generalizability of DNNs for detecting the operational context followed by the context-specific errors. While DNNs have been widely successful across many domains, slight perturbations in the input data brought about by the noise in the environment \cite{zheng2016improving} or attacks \cite{goodfellow2014explaining} can lead them to misclassify with high confidence. However, most of the proposed adversarial examples on DNNs target image-based classification systems. Our safety monitoring system is based on kinematics samples and we only use computer vision for orthogonal labeling of failures. A further robustness analysis and design of our ML-based safety monitor against accidental and malicious perturbations is the subject of future work. 

In addition, the performance of supervised learning models heavily depends on the accurate labeling of the operational context, or surgical gestures, and the context-specific anomalies, or erroneous gestures. Our labeling of the erroneous gestures for the JIGSAWS dataset was based on the human annotations of the corresponding videos. We labeled any gesture that had an occurrence of an anomaly as erroneous even if the error did not occur at the beginning of the gesture. Future work will focus on automated labeling of trajectory data from real surgical tasks (similar to our automated labeling of Block Transfer task on RAVEN II robot using video data) for more precise localization of errors. 


\section{CONCLUSION}
We presented an end-to-end safety monitoring system for real-time context-aware identification of erroneous gestures in robotic surgery. Our preliminary results show the promise of our kinematics-only based solution in timely and accurate detection of unsafe events, even when the vision data might not be available or be sub-optimal. Our experimental results validate the need for context-aware monitoring, while also suggesting that some surgical gestures have similar error-patterns and can potentially be better monitored together as a sequence. Our results also show the potential for early detection and prevention of these unsafe events, which could be further enhanced by having access to larger training datasets and extending the semantics of context using vision or other sensing modalities. Future work will focus on the generalization of our solution to a wider set of realistic surgical gestures and tasks with a larger number of trials. We also plan to further improve the accuracy and timeliness of our safety monitoring system to enable successful prevention of safety-critical events during surgery.






\bibliographystyle{IEEEtran}
\bibliography{references.bib}
\end{document}